# MAAPO: an innovative membrane algorithm based on artificial protozoa optimizer for multilevel threshold image segmentation

**Xiaopeng Wang[1] · Václav Snášel[1] · Seyedali Mirjalili[2] · Jeng-Shyang Pan[3]**

**Abstract**
This paper proposes a novel membrane algorithm based on artificial protozoa optimizer (MAAPO) for global optimization problems. The artificial protozoa optimizer (APO) is adopted as the base meta-heuristic algorithm due to its novelty and competitive performance. MAAPO integrates two key innovations: (1) a membrane computing (MC) framework that introduces a parallel distributed paradigm to improve population diversity and search dynamics, and (2) an enhanced autotrophic model within APO that uses a roulette-based fitness-distance balance (RFDB) mechanism for adaptive reference point selection. These strategies collectively enhance the algorithm's exploration-exploitation balance and global search capabilities. To validate its performance, MAAPO is tested against 12 advanced algorithms on the CEC2017 test suite, and further applied to the multilevel thresholding image segmentation problem using Otsu and Kapur entropy as objective functions. The quality of segmented images is assessed using peak signal-to-noise ratio (PSNR), structural similarity index (SSIM), and feature similarity index (FSIM) metrics. Experimental results demonstrate that MAAPO outperforms its counterparts, delivering superior segmentation quality. This research on MAAPO contributes an effective enhancement strategy to meta-heuristic algorithms and introduces a novel, highly applicable approach for complex image segmentation tasks. The source codes of MAAPO are publicly available at https://ww2.mathworks.cn/matlabcentral/fileexchange/181534-maapo.

**Keywords** Meta-heuristic algorithm · Artificial protozoa optimizer · Membrane computing · Fitness-distance balance · Image segmentation

✉ Xiaopeng Wang
xiaopeng.wang@vsb.cz

✉ Václav Snášel
vaclav.snasel@vsb.cz

[1] Faculty of Electrical Engineering and Computer Science, VŠB-Technical University of Ostrava, Ostrava, Czech Republic

[2] Centre for Artificial Intelligence Research and Optimisation, Torrens University Australia, Brisbane, Australia

[3] School of Artificial Intelligence, Nanjing University of Information Science and Technology, Nanjing, China

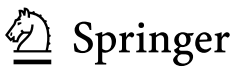

## 1 Introduction

Nowadays, digital images are integral to modern electronic devices and embedded systems. They play a critical role in a wide range of applications, including security monitoring, autonomous vehicles, robotic vision, medical imaging, smart agriculture, environmental monitoring, traffic management, and digital media. Their widespread adoption highlights their importance in enabling intelligent, data-driven technologies (Oliva et al. 2019; Kaur and Kaur 2014). Image processing technology employs computer algorithms to process, analyze, and interpret images for various applications. An essential component of this process, image segmentation, focuses on identifying and extracting meaningful objects, regions, and boundaries to support deeper analysis. Traditional segmentation methods like edge-based, region-based, and clustering-based approaches are widely used, but each comes with notable limitations. Edge-based segmentation methods are sensitive to noise and gradual intensity variations, making them ineffective for images with low contrast since detecting robust edges is challenging. Region-based segmentation methods tend to be computationally heavy, especially for high-resolution images. It is also sensitive to initial conditions, and randomly selected initial seeds cause the algorithm to get trapped in local optima, leading to under-segmentation (merging different regions) or over-segmentation (splitting homogeneous regions). Clustering-based segmentation methods struggle with noise and outliers, which can degrade segmentation accuracy. Moreover, they are highly sensitive to initial conditions; for instance, K-means requires predefining the number of clusters. In practical scenarios, determining the optimal number of clusters beforehand is difficult. Additionally, these methods can be computationally expensive.

With the growing demand for precise visual understanding in real-world scenarios, there is strong motivation to develop more advanced segmentation techniques. Recent advances in threshold-based methods have attracted considerable attention due to their computational simplicity, robustness, and effectiveness. Thresholding methods assign pixels to different classes based on gray-level intensity. According to the number of thresholds applied, they are typically categorized into bilevel and multilevel thresholding methods. Bilevel thresholding segments an image into two classes using a single threshold value. Pixels with intensity values below the threshold are classified as background, while those above are identified as foreground or objects. In contrast, multilevel thresholding employs two or more thresholds to partition the image into multiple regions. In general, increasing the number of thresholds can improve segmentation quality. However, as the number of thresholds rises, the problem becomes more complex and computationally demanding. Nevertheless, multilevel thresholding remains a preferred approach in practical applications due to its effectiveness.

Meta-heuristic algorithms are powerful optimization techniques that are inspired by natural phenomena. They possess several advantages, such as strong global search capabilities, easy implementation, robustness, and flexibility. Consequently, they have been extensively applied in diverse applications such as engineering design (Zhang et al. 2024), path planning (Liu et al. 2023; Dai et al. 2023), photovoltaic models (Wang et al. 2021; Li et al. 2022), feature selection (Wang et al. 2024; Hu et al. 2020), traveling salesman problem (Panwar and Deep 2021; Changdar et al. 2023), energy vehicle dispatch (Liang et al. 2019), neural network training (Abd Elaziz et al. 2021), and wireless sensor networks (Mann and Singh 2019; Del-Valle-Soto et al. 2023). Finding ideal thresholds is essential for effective image segmentation, and meta-heuristic algorithms have proven to be valuable tools in achieving this goal. Yin (Yin 2007) introduced a segmentation strategy in 2007 that combined particle swarm optimization with the minimum cross-entropy function. The study by Oliva et al. (2014), published in 2014, developed a multilevel thresholding method that utilized electromagnetism optimization to optimize both Otsu's criterion and Kapur's entropy. Sarkar et al. (2015) proposed a multilevel thresholding framework in 2015, which used the differential evolution algorithm to optimize the minimum cross-entropy objective. In 2020, Yang and Wu (2020) proposed an improved particle swarm optimization algorithm for image segmentation tasks. An exchange market algorithm was proposed by Sathya et al. (2021) in 2021 to optimize the objective functions of Otsu, Kapur, and minimum cross-entropy methods. In 2022, Chen et al. (2022) developed a multi-strategy shuffled frog leaping algorithm and successfully applied it to breast invasive ductal carcinoma segmentation. In 2024, Bei et al. (2024) developed an improved slime mould algorithm called SMA-MLS that used Kapur entropy as its fitness function. Most recently, in 2025, Abd Elaziz et al. (2025) presented an improved planet optimization algorithm specifically designed for polyp image segmentation. In addition, various image segmentation techniques have been developed based on different meta-heuristic algorithms. These include the genetic algorithm (Yin 1999; Canales et al. 2024), multi-verse optimizer (Kandhway and Bhandari 2019; Wang et al. 2020), grey wolf optimizer (Khairuzzaman and Chaudhury 2017; Yu and Wu 2022), teaching-learning-based optimization (Jiang et al. 2022; Wu et al. 2020), marine predators algorithm (Abd Elaziz et al. 2021; Abdel-Basset et al. 2022), gravitational search algorithm (Jiao and Pan 2019), and poplar optimization algorithm (Chen et al. 2022).

Despite the development of various algorithms for image segmentation, it is still challenging to achieve accurate results and avoid convergence to local optima. Segmenting images with unknown or complex attributes remains difficult, and stable thresholds are hard to obtain. Multidimensional thresholding is effective for fine and accurate segmentation, but increasing the number of thresholds or image information dimensions leads to exponential growth in time complexity. This demands better convergence, higher accuracy, and stronger capabilities to avoid local optima. In this study, we introduce a new membrane algorithm based on artificial protozoa optimizer (MAAPO) to effectively solve the multilevel thresholding image segmentation problem. The basic artificial protozoa optimizer (APO) (Wang et al. 2024) is a newly published meta-heuristic algorithm. Its search operator simulates the foraging, dormancy, and reproduction behaviors of protozoa. APO provides better solutions than similar algorithms in most global optimization problems. However, it may still get stuck in local optima and converge prematurely in complex optimization situations. Therefore, we integrate membrane computing (MC) and roulette fitness-distance balance (RFDB) mechanisms to improve the performance of the algorithm. In MAAPO, MC establishes a parallel and distributed framework. Its separating'merging principle enables dynamic adjustment of each membrane's population size during the optimization process, thereby promoting a more effective balance between exploration and exploitation. Additionally, the RFDB selection strategy identifies promising reference points based on both fitness values and their relative distances within the search space. These reference points serve as guidance anchors, enabling the population to move toward more optimal regions and enhancing the overall convergence behavior of the algorithm. The performance of the proposed MAAPO algorithm is first validated on the CEC 2017 test suite. Furthermore, the algorithm is integrated with Otsu and Kapur entropy methods to solve multilevel thresholding image segmentation tasks. The segmented images are evaluated using peak signal-to-noise ratio (PSNR) (Wang et al. 2004), structural similarity index (SSIM) (Ndajah et al. 2010), and feature similarity index (FSIM) (Zhang et al. 2011). The results demonstrate the proposed algorithm's strong performance in image segmentation. The main contributions of this study can be summarized as follows:

1. This study introduces MAAPO, a new membrane algorithm derived from the artificial protozoa optimizer.
2. The algorithm incorporates membrane computing and roulette fitness-distance balance mechanisms to enhance exploration and exploitation abilities.
3. A comprehensive evaluation under the CEC2017 test suite demonstrates that MAAPO surpasses 12 state-of-the-art algorithms.
4. In multilevel thresholding image segmentation tasks, MAAPO achieves more competitive results than its counterparts.

The remainder of this paper is organized as follows. Section 2 presents the preliminaries, including an overview of the APO algorithm and membrane computing. Section 3 details the proposed MAAPO algorithm along with the roulette fitness-distance balance (RFDB) selection mechanism. Section 4 conducts an empirical evaluation of MAAPO using the CEC2017 test suite. Section 5 applies MAAPO to address the multilevel threshold image segmentation problem. Finally, Sect. 6 concludes this study and outlines potential future research.

## 2 Preliminaries

This section presents the underlying concepts of the proposed algorithm. It first describes the standard APO algorithm, then introduces membrane computing as a supporting framework.

### 2.1 Artificial protozoa optimizer

The artificial protozoa optimizer mimics the foraging, dormancy, and reproduction of protozoa in nature. Protozoa can exhibit two types of foraging behaviors: autotrophic and heterotrophic modes. In the light, protozoa perform autotrophic foraging similar to plants, producing carbohydrates through photosynthesis. In the dark, protozoa behave like animals in heterotrophic foraging, absorbing organic matter through phagocytosis. In adverse environments, protozoa slow down their metabolism and enter a dormant state to reduce their dependence on the surrounding environment. At the appropriate age, protozoa will reproduce by binary fission, producing two offspring. The search operators of the APO algorithm are developed by mimicking the behavioral patterns of protozoa, as described below.

- Autotrophic foraging

$$X_i^{new} = X_i + f \cdot (X_j - X_i + \frac{1}{np} \cdot \sum_{k=1}^{np} w_a \cdot (X_{k-} - X_{k+})) \odot M_f \tag{1}$$

$$f = rand \cdot (1 + \cos(\frac{iter}{iter_{max}} \cdot \pi)) \tag{2}$$

$$w_a = e^{-\left|\frac{f(X_{k-})}{f(X_{k+})+eps}\right|} \tag{3}$$

$$M_f[di] = \begin{cases} 1, & \text{if } di \text{ is in } randperm(dim, \lceil dim \cdot \frac{i}{ps} \rceil) \\ 0, & \text{otherwise} \end{cases} \tag{4}$$

where:

- $X_i^{new}$ and $X_i$ are the new and current positions of the *i*th protozoan, respectively.
- $X_j$ is the randomly selected *j*th protozoan.
- In the *k*th paired neighbor, $X_{k-}$ is a randomly chosen protozoan with a rank index less than *i*.
- In the *k*th paired neighbor, $X_{k+}$ is a randomly chosen protozoan with a rank index greater than *i*.
- *f* is a foraging factor. *rand* is a random number in the (0,1). *iter* is the current iteration, and $iter_{max}$ is maximum iteration.
- *np* indicates the number of neighboring pairs and is set to 1.
- $w_a$ is a weight factor for autotrophic foraging. $f(X_{k-})$ is the fitness value of $X_{k-}$, and $f(X_{k+})$ is the fitness value of $X_{k+}$. *eps* is a very small number equal to 2.2204e−16.
- $\odot$ is the Hadamard product.

- $M_f$ is the foraging mapping vector; its size is $(1 \times dim)$ and each element is 0 or 1. *di* is the dimension index. *dim* is the number of decision variables. *ps* is the population size. *randperm*(*n*, *l*) provides a row vector of *l* different integers, randomly chosen from 1 to *n*.Heterotrophic foraging

$$X_i^{new} = X_i + f \cdot (X_{near} - X_i + \frac{1}{np} \cdot \sum_{k=1}^{np} w_h \cdot (X_{i-k} - X_{i+k})) \odot M_f \tag{5}$$

$$X_{near} = (1 \pm Rand \cdot (1 - \frac{iter}{iter_{max}})) \odot X_i \tag{6}$$

$$w_h = e^{-\left|\frac{f(X_{i-k})}{f(X_{i+k})+eps}\right|} \tag{7}$$

where:

- $X_{near}$ is a nearby position of the *i*th protozoan. "±" indicates that $X_{near}$ can be selected from different directions. *Rand* is a random vector, its size is $(1 \times dim)$ and each element is a random number in the (0,1).

- $w_h$ is the weight factor for heterotrophic foraging. $f(X_{i-k})$ is the fitness values of $X_{i-k}$, and $f(X_{i+k})$ is the fitness values of $X_{i+k}$.Dormancy

$$X_i^{new} = X_{min} + Rand \odot (X_{max} - X_{min}) \tag{8}$$

where:

- $X_{min}$ is the lower-bound vector.

- $X_{max}$ is the upper-bound vector.Reproduction

$$X_i^{new} = X_i \pm rand \cdot (X_{min} + Rand \odot (X_{max} - X_{min})) \odot M_r \tag{9}$$

$$M_r[di] = \begin{cases} 1, & \text{if } di \text{ is in } randperm(dim, \lceil dim \cdot rand \rceil) \\ 0, & \text{otherwise} \end{cases} \tag{10}$$

where:

- "±" indicates that the perturbation may be either forward or backward.
- $M_r$ is a mapping vector for reproduction; its size is $(1 \times dim)$ and each element is 0 or 1.

To complete the algorithm design, three parameters controlling the selection of search operators are introduced below.

$$pf = pf_{max} \cdot rand \tag{11}$$

$$p_{ah} = \frac{1}{2} \cdot (1 + \cos(\frac{iter}{iter_{max}} \cdot \pi)) \tag{12}$$

$$p_{dr} = \frac{1}{2} \cdot (1 + \cos((1 - \frac{i}{ps}) \cdot \pi)) \tag{13}$$

where:

- *pf* is the proportion fraction of the protozoan population in dormancy and reproductive states. $pf_{max}$ is the maximum *pf* and is set to 0.1.
- $p_{ah}$ is a probability parameter that determines whether the protozoan is in autotrophic or heterotrophic foraging state.
- $p_{dr}$ is a probability parameter that determines whether the protozoan is in a dormant or reproductive state.

For a clearer understanding, Fig. 1 illustrates the relationship between exploration/exploitation and the four search operations. Autotrophic foraging and dormancy support exploration, whereas heterotrophic foraging and reproduction support exploitation. The value of $p_{ah}$ gradually decreases over iterations, thereby supporting the transition of protozoa from autotrophic to heterotrophic foraging. For dormancy and reproduction behaviors, the $p_{dr}$ parameter is affected by the protozoan ranking index (protozoa are ranked based on fitness values in each iteration). Among them, superior protozoa tend to reproduce, while inferior protozoa are more likely to enter dormancy. This enables exploration not only during the initial iterations but also extends to the final iterations, alleviating local optimum stagnation.

**Fig. 1** Exploration and exploitation

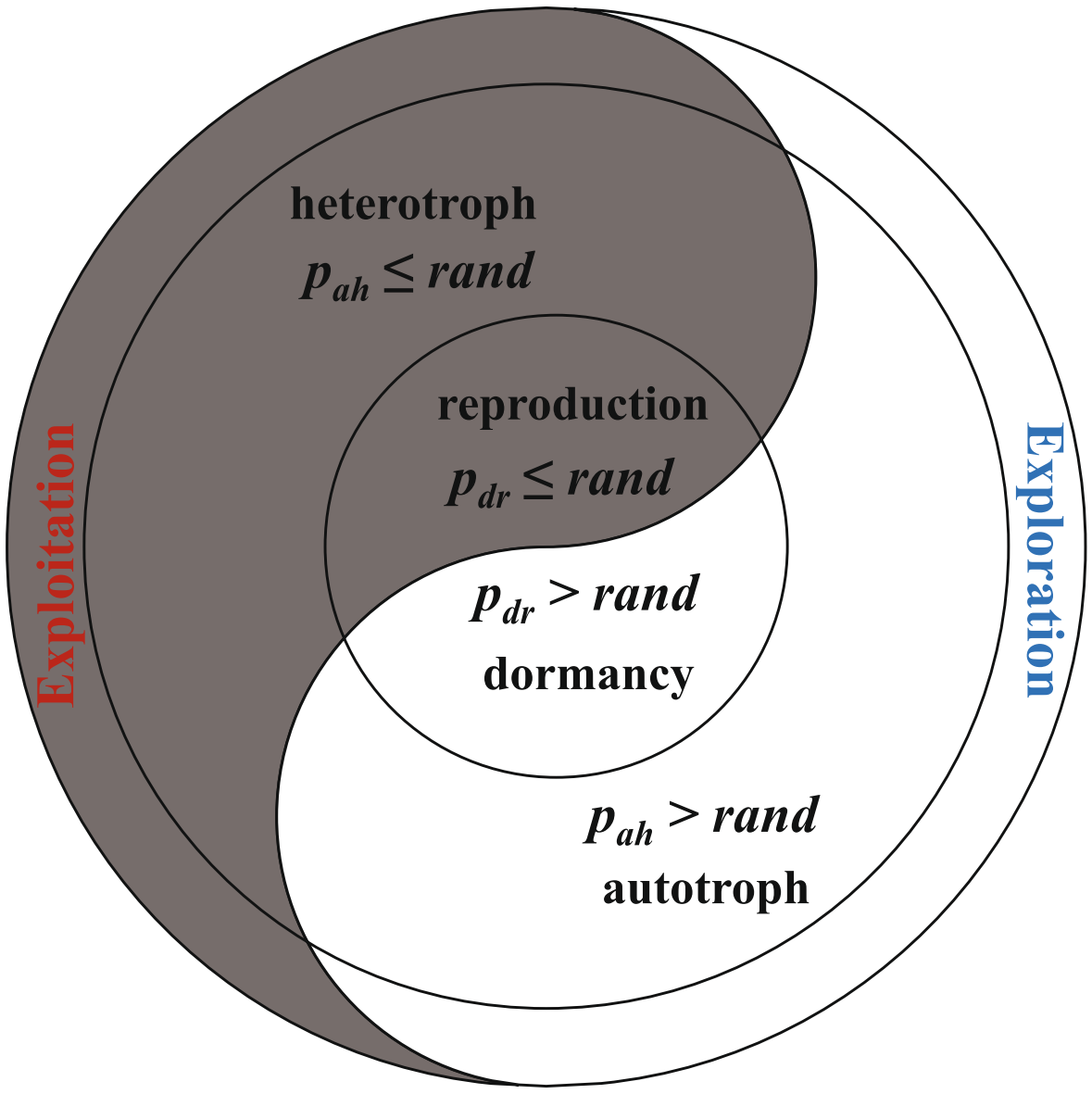

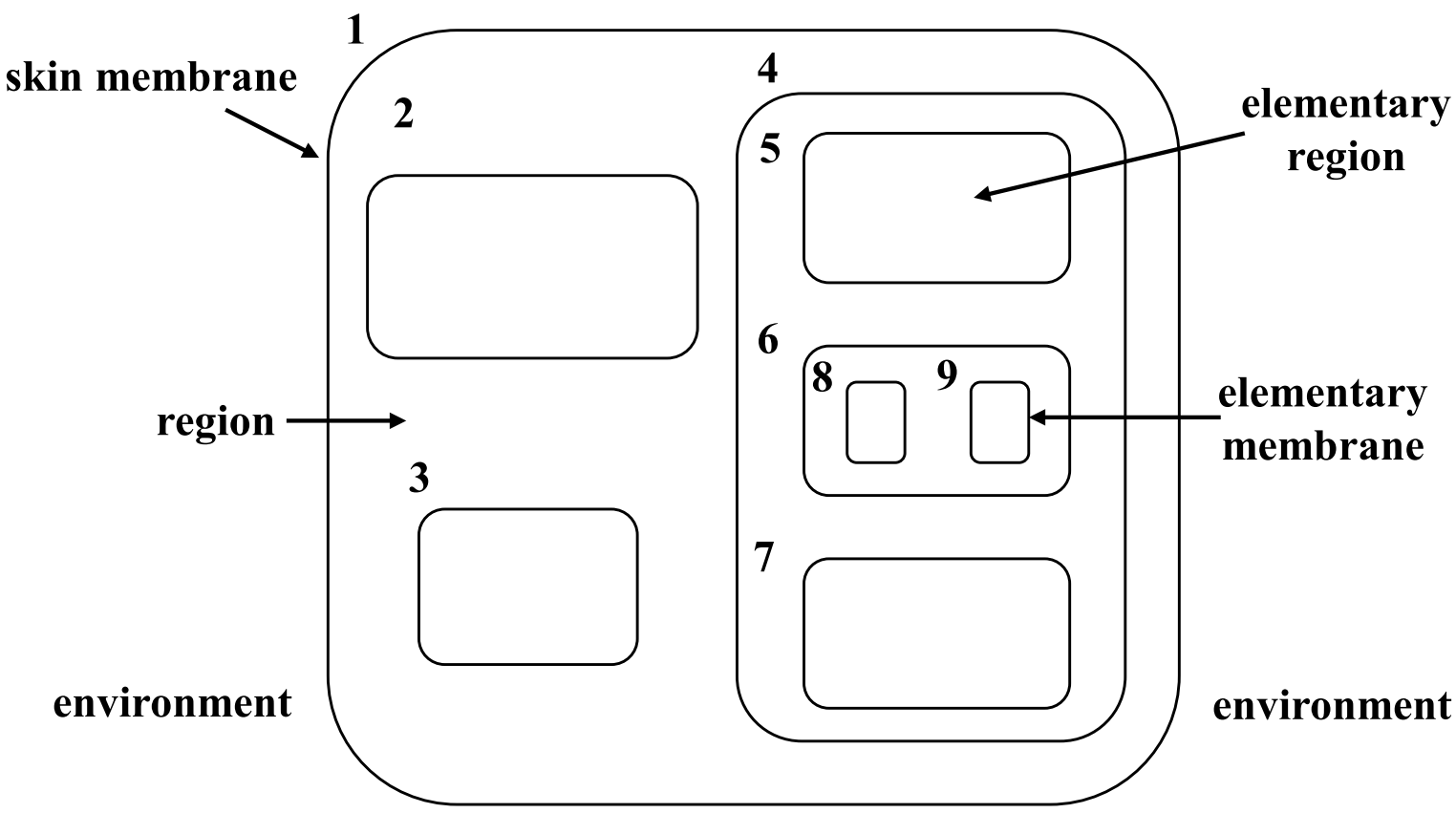

**Fig. 2** The membrane structure of cell-like system

## 2.2 Membrane computing

Membrane computing (MC) simulates the dynamics of molecular evolution and intermolecular communication in biological cells. ItÂ provides a parallel distributed system with objects, rules, and structure, known as the membrane or P-system. The membrane system is classified into cell-like, tissue-like, and neural-like systems. In this study, the cell-like system is used, and Fig. 2 (Păun and Rozenberg 2002) depicts a membrane structure. The skin membrane, the outermost membrane, divides the system from the environment. The elementary membrane is the innermost membrane without other submembranes. The objects contained in each membrane will evolve based on updated rules (Cheng et al. 2011; Alsalibi et al. 2022).

The definition of a cell-like membrane system with $m$ membranes is as follows:

$$\Pi = (O, \mu, \omega_i, M_i, R_i, \rho_i, i_o) \tag{14}$$

where:

- $O$ is an alphabet of objects.
- $\mu$ deontes a membrane structure with $m$ membranes, and $m$ is the degree of the system.
- $\omega_i$ is a string, representing the objects linked to the $i$th membrane $(1 \leq i \leq m)$.
- $M_i$ is the $i$th membrane's initial objects.
- $R_i$ is the $i$th membrane's updated rules.
- $\rho_i$ is the rule priority of the $i$th membrane.
- $i_o$ is the output membrane label.

The following lists the five primary methods for the updated rules: Evolution rules: A variety of evolution operators are used to modify the objects. In-communication: The objects are transmitted into the membrane. Out-communication: The objects are transmitted out from the membrane. Merging rules: Two or more membranes combine to form a single membrane. Separation rules: Two or more membranes separate from a membrane.

## 3 Proposed MAAPO algorithm

The membrane algorithm is a hybrid approach that combines the distributed parallelism of membrane systems with the search efficiency of meta-heuristic algorithms. Membrane systems provide a solid theoretical basis for parallel computation, while meta-heuristics offer powerful yet flexible optimization strategies. Since its introduction, the membrane algorithm has attracted significant research interest. This section introduces the proposed MAAPO algorithm, including its framework, the APO variant with fitness-distance balance (FDB), and computational complexity.

### 3.1 Framework of MAAPO

The membrane system of MAAPO is designed with separation and merging principles to facilitate population evolution. Each solution is regarded as an object in the elementary membrane. In each iteration, when the multidimensional volumes ($mVOL$), a diversity metric of the population, is smaller than the threshold $\varepsilon$, all updated solutions are contained in an elementary membrane. Alternatively, the separating-merging operation is activated, and all solutions are randomly divided into several membranes, updated separately, and then merged. The best result is output when the predefined termination is reached. The framework of MAAPO is shown in Fig. 3. The hyperparameters involved are $mVOL$, $\varepsilon$, and $m$ (the number of separated membranes).

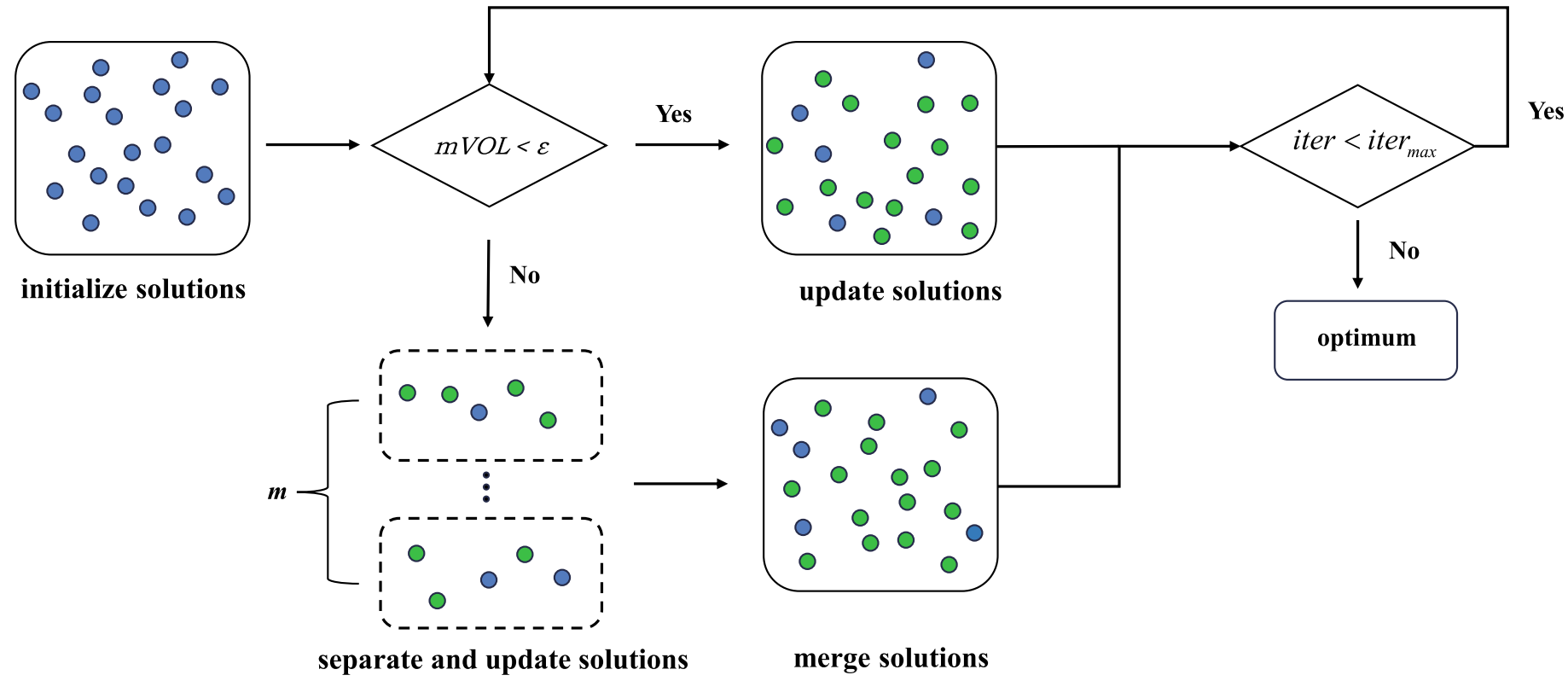

**Fig. 3** The framework of the proposed MAAPO algorithm

A modified version of Osuna-Enciso et al. (2022) is introduced, called the $mVOL$ metric, to detect the spatial distribution of solutions within the population. This diversity metric consists of two hypervolumes. The first is called $V_{pop}$, and it shows the current distribution of the population in the iteration. The other is $V_{lim}$, which represents the limit of the search space. $mVOL$ is computed using Eq. (15). This work sets $\varepsilon$ to 0.0001 and calculates the parameter $m$ using Eq. (18). The parameter setting is studied in the experimental part.

$$mVOL = \left(\frac{V_{pop}}{V_{lim}}\right)^{1/dim} \tag{15}$$

$$V_{pop} = \prod_{i=1}^{dim} |\max(dim_i) - \min(dim_i)| \tag{16}$$

$$V_{lim} = \prod_{i=1}^{dim} |u_i - l_i| \tag{17}$$

where:

- $dim$ is the number of decision variables.
- $\max(dim_i)$ is the maximum value of the $ith$ dimension.
- $\min(dim_i)$ is the minimum value of the $ith$ dimension.
- $u_i$ and $l_i$ are the upper and lower bounds of the $ith$ dimension respectively.

$$m = m_{min} + \lceil (m_{max} - m_{min}) \cdot rand \rceil \tag{18}$$

where:

- $m_{min}$ is set to 1.
- $m_{max}$ is set to 4.

– $\lceil \cdot \rceil$ is the ceiling function.

### 3.2 APO variant with FDB

The selection mechanism is a critical component in the design of meta-heuristic algorithms. Selecting promising individuals from the population to engage in search operations can significantly enhance the efficiency of the search process. Kahraman and Gedikli proposed a selection method called fitness-distance balance (FDB) in 2020 (Kahraman et al. 2020). This method takes into account distance and fitness value factors, which is beneficial to enhancing the diversity of the population and alleviating premature convergence. The following describes FDB and its variants, dynamic fitness-distance balance (DFDB) (Kahraman et al. 2022), and roulette fitness-distance balance (RFDB) (Bakır 2024).

- FDB

$$S(X_i) = w \times norm(f(X_i)) + (1 - w) \times norm(d(X_i)) \tag{19}$$

where:

  - $S(X_i)$ is the FDB score of the *ith* individual, and the one with the highest score will be selected.
  - *w* is the weight coefficient. In this study, it is set to 0.5.
  - $norm(f(X_i))$ is the normalized fitness value of the *ith*.
  - $norm(d(X_i))$ is the normalized Euclidean distance between the *ith* individual and the best ones,

- DFDB

The DFDB selection dynamically modifies the parameter *w*. As shown in Eq. (20), *w* is small in the early stage, and the DFDB method will select a solution far away from the best solution to contribute to the exploration ability. As *w* increases with iterations, the fitness value factor has a greater impact on the DFDB score. This will intensify the search around quality solutions and contribute to the exploitation ability of the algorithm.

$$w = \frac{iter}{iter_{max}} \cdot (1 - w_{min}) + w_{min} \tag{20}$$

where:

- $w_{min}$ is the minimum weight parameter. In this study, it is set to 0.2.RFDB

The RFDB selection is a combination of roulette wheel selection and FDB. Compared with FDB selection, RFDB not only considers greed but also focuses on randomness. This is a probability-based method that favors selecting individuals with high FDB scores. The probability of the *ith* individual being selected is $S(X_i)/\sum_{i=1}^{ps} S(X_i)$.

In the APO variant, autotrophic foraging is modified using RFDB for the selection of $X_j$, and Eq. (1) is updated accordingly as follows:

$$X_i^{new} = X_i + f \cdot (X_{RFDB} - X_i + \frac{1}{np} \cdot \sum_{k=1}^{np} w_a \cdot (X_{k-} - X_{k+})) \odot M_f \tag{21}$$

Figure 3 illustrates the framework of MAAPO. The process steps are outlined in the following. Step 1: Initialize the population using random sampling, and evaluate the fitness values. Step 2: Calculate the diversity metric *mVOL* to identify the membrane structure. Step 3: The APO variant with RFDB updates the solutions in each elementary membrane. If $mVOL < \varepsilon$, all solutions are contained in one membrane structure. Otherwise, the separating-merging operation is used. Step 4: Update the global optimum. When the predefined termination condition is satisfied, output the optimum; if not, return to step 2. The pseudo code of the proposed MAAPO is given in Algorithm 1.

```
Input: ps, dim, iter_max, np, pf_max, ε, and m.
Output: The global optima X_gbest and f(X_gbest).
1:  while iter < iter_max do
2:      Calculate mVOL using Eq. (15).
3:      if mVOL < ε then
4:          for each solution within one membrane do
5:              Update X_i using Eq. (21), Eq. (5), Eq. (8), and Eq. (9).
6:          end for
7:      else
8:          The m membranes are created by randomly separating the solutions inside one membrane.
9:          for each solution within m membranes do
10:             Update the solution using Eq. (21), Eq. (5), Eq. (8), and Eq. (9).
11:         end for
12:         A membrane is created by merging m membranes.
13:     end if
14:     X_gbest = opt{X_i}.
15:     iter ← iter + 1.
16: end while
```

**Algorithm 1** The pseudo code of MAAPO

## 3.3 Computational complexity

The proposed MAAPO algorithm includes the sorting, search operators, and fitness function evaluation from the APO algorithm; the calculation of the *mVOL* metric, and the RFDB selection. In each iteration, the computational complexity of the APO algorithm is $O(ps \cdot (\log(ps) + dim + f(\cdot)))$, the calculation of *mVOL* metric is $O(ps \cdot dim)$, and the RFDB selection is $O(ps^2 + ps \cdot dim)$. Therefore, the total computational complexity is $O(iter_{max} \cdot ps \cdot (ps + dim + f(\cdot)))$.

## 4 Empirical studies

This section evaluates the performance of MAAPO under the CEC2017 test suite (Wu et al. 2017). The benchmark consists of 30 functions, covering unimodal, multimodal, hybrid, and composition categories. The Wilcoxon signed-rank test is used at a 5% significance level to assess differences between two algorithms, while the Friedman test is used to compare multiple algorithms. The experimental simulations were conducted on a personal laptop running Windows 11, equipped with 16 GB RAM and an Intel(R) Core(TM) i7-8750 H CPU @ 2.20 GHz 2.21 GHz.

### 4.1 Experiment settings

The simulation follows the definition standard of CEC2017. The dimension is 10, the maximum number of fitness evaluations is 100,000, the population size is 100, and each algorithm is tested 51 times. The global minimum of each function is shifted to "0" and error values less than $10^{-8}$ are considered "0." For the MAAPO algorithm, the parameter settings are as follows: *np* = 1, $pf_{max}$ = 0.1, $\varepsilon$ = 0.0001, $m_{min}$ = 1, and $m_{max}$ = 4. The compared algorithms include artificial protozoa optimizer (APO), grey wolf optimizer (GWO) (Mirjalili et al. 2014), improved grey wolf optimizer (IGWO) (Nadimi-Shahraki et al. 2021), particle swarm optimization (PSO) (Kennedy and Eberhart 1995), parallel particle swarm optimization (PPSO) (Chu et al. 2005), differential evolution (DE) (Price et al. 2006), artificial rabbits optimization (ARO) (Wang et al. 2022), five phases algorithm (FPA) (Wang et al. 2023), gannet optimization algorithm (GOA) (Pan et al. 2022), membrane algorithm based on quasi-affine transformation evolution (MAQUATRE) (Wang et al. 2023), snow geese algorithm (SGA) (Tian et al. 2024), and animated oat optimizer (AOO) (Wang et al. 2025). Table 1 lists the parameter configurations along with the publication years of the algorithms.

**Table 1** Algorithm parameter configurations

| Algorithm | Parameter settings | Year |
|---|---|---|
| MAAPO | $np = 1, pf_{max} = 0.1,$ $\varepsilon = 0.0001,$ $m_{min} = 1, m_{max} = 4$ | – |
| APO | $np = 1, pf_{max} = 0.1$ | 2024 |
| GWO | $a = [2, 0]$ | 2014 |
| IGWO | $a = [2, 0]$ | 2021 |
| PSO | $v_{min} = -10, v_{max} = 10,$ $w = [0.9, 0.4], c_1, c_2 = 2$ | 1995 |
| PPSO | $v_{min} = -10, v_{max} = 10,$ $w = [0.9, 0.4], c_1, c_2 = 2,$ $group = 4, R = 20$ | 2005 |
| DE | $f = 0.7, cr = 0.1$ | 1995 |
| ARO | No special parameters | 2022 |
| FPA | No special parameters | 2023 |
| GOA | $r = 0.5, q = 0.5, c = 0.2$ | 2022 |
| MAQUATRE | $f = 0.7$ | 2023 |
| SGA | $\rho = 1.29$ | 2024 |
| AOO | $\beta = 1.5$ | 2025 |

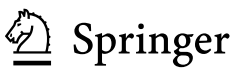

**Table 2** Performance comparison of APO and three variants

| Algorithm | APO vs. | FDB-APO | DFDB-APO | RFDB-APO |
|---|---|---|---|---|
| +/=/− | – | 16/10/4 | 17/5/8 | 0/27/3 |

**Table 3** Ranking results of MAAPO under varying $\varepsilon$ settings

| $\varepsilon$ | 1 | 0.1 | 0.01 | 0.001 | 0.0001 | 0.00001 | 0 |
|---|---|---|---|---|---|---|---|
| Average ranking | 4.85 | 4.13 | 3.80 | 3.63 | **3.60** | 4.30 | 3.68 |

The best results are shown in bold

**Table 4** Ranking results of MAAPO under varying *m* settings

| *m* | 1 | 2 | 4 | 8 | $m_{max}$=4 | $m_{max}$=8 |
|---|---|---|---|---|---|---|
| Average ranking | 4.12 | 3.45 | 3.58 | 3.52 | **3.08** | 3.25 |

The best results are shown in bold

## 4.2 Analysis of APO variants

This subsection introduces three variants of the APO algorithm, namely FDB-APO, DFDB-APO, and RFDB-APO. These variants differ in how they select $X_j$ during autotrophic foraging, using the FDB, DFDB, and RFDB strategies, respectively. The variants are tested on CEC2017 and compared with the original APO. Table 2 provides an overview of the simulation. The symbols "+", "=", and "−" denote that APO performs better than, equal to, or worse than the compared variants, respectively. In the Wilcoxon signed-rank test method [59], APO may beat FDB-APO 16 times, draw 10 times, and lose 4 times; beat DFDB-APO 17 times, draw 5 times, and lose 8 times; but compared with RFDB-APO, APO has 0 wins, 27 draws, and 3 losses. The results indicate that RFDB-APO is the best variant. This is because APO's autotrophic foraging model controls its exploration process, whereas FDB and DFDB deterministically select a single reference point in each iteration, which limits the algorithm's ability to explore effectively. However, RFDB incorporates the roulette wheel selection to enable the selection of diverse reference points in each iteration. Therefore, the RFDB-APO algorithm is employed to update the population within each elementary membrane of the MAAPO framework.

## 4.3 Hyperparameter settings for the MAAPO framework

In this subsection, various $\varepsilon$ and *m* parameters are tested to select competitive values. Experiments are conducted with multiple $\varepsilon$ settings while fixing *m* at 2. The Friedman test method (Derrac et al. 2011) ranks the mean fitness functions under each parameter setting. The results of 30 functions are summarized in Table 3. $\varepsilon = 1$ denotes that the algorithm performs only one membrane operation in each generation, while $\varepsilon = 0$ means that only separating-merging membrane operations are performed. The separating-merging operation frequency is given by $\varepsilon$ in (0, 1). The results confirm that the most competitive one is $\varepsilon$ set to 0.0001, with an average rank of 3.60. The worst one is $\varepsilon$ set to 1, with an average rank of 4.85. This indicates that the separating-merging membrane principle can boost the algorithm's performance.

Furthermore, the impact of different *m* values was examined, with $\varepsilon$ set to 0.0001. Six different scenarios were evaluated by setting *m* to 1, 2, 4, 8, and $m_{max}$ to 4 and 8. Table 4 presents the ranking results for different *m* values. The setting $m_{max}$ = 4 achieves the best performance with an average ranking of 3.08, whereas the setting *m* = 1 performs the worst, with an average ranking of 4.12. It can be observed that dividing the population into distinct membranes enhances the algorithm's performance. Additionally, the dynamic membrane number performs better than the static membrane number. Therefore, it is recommended to set $\varepsilon$ to 0.0001 and $m_{max}$ to 4 in the MAAPO algorithm.

### 4.4 Ablation experiment

Ablation studies were designed to verify the contribution of each strategy. Specifically, MS-APO refers to APO combined solely with the membrane system strategy, while RFDB-APO denotes APO integrated only with the RFDB selection strategy. The performance comparisons are presented in Table 5. The Wilcoxon signed-rank test results reveal that MAAPO delivers better performance than MS-APO and RFDB-APO in 5 out of 30 test functions, exhibits no statistically significant difference in 25 cases, and is not inferior in any. In comparison with APO, MAAPO demonstrates improvements in 8 functions, comparable performance in 22, and no degradation in any case. These findings validate that each added strategy enhances the performance of APO. Both the RFDB selection method and the membrane system are essential to MAAPO's success, as removing either leads to performance degradation. Therefore, integrating RFDB with the membrane system synergistically enhances APO's overall optimization capability, enabling MAAPO to achieve superior performance on the test functions.

### 4.5 Comparative study of MAAPO and state-of-the-art algorithms

The proposed MAAPO is evaluated against 12 state-of-the-art algorithms. Table 6 presents the simulation results. "Mean" denotes the average fitness value, and "Std" denotes the standard deviation. The best statistical result for each function is bolded. The Wilcoxon signed-rank test indicates that the proposed MAAPO performs significantly better, comparable to, or worse than the compared algorithms, denoted by the symbols "+", "=", and "–", respectively. The Wilcoxon signed-rank test results indicate that MAAPO achieves a superior or comparable performance in most cases. Notably, MAAPO records 30 wins and no losses against both GWO and SGA, demonstrating dominant performance. Against APO, MAAPO achieves 8 wins, 22 draws, and no losses, suggesting improved behavior. MAAPO also shows strong results against IGWO, PSO, PPSO, and GOA, with only a small number of losses. Compared to DE, MAAPO achieves 19 wins, 7 draws, and 4 losses. In the case of ARO, MAAPO secures 21 wins, 7 draws, and 2 losses. In contrast to FPA, MAAPO wins 22 times, draws 6 times, and loses 2 times. When compared with MAQUATRE, the outcomes are evenly split, with 10 wins, 10 draws, and 10 losses. In contrast, MAAPO achieves a clear advantage over AOO, with 25 wins and no losses. MAAPO achieved the best results in 14 of the 30 tested functions, including F2, F5-F9, F13, F14, F16, F18, F21, F23, F27, and F29. According to the Friedman test, the rankings are as follows: MAAPO ranks first with a score of 3.35, followed by MAQUATRE with 3.37, APO with 3.55, DE with 4.43, ARO with 5.25, FPA with 6.90, PPSO with 7.10, IGWO with 7.32, GOA with 7.95, AOO with

**Table 5** Ablation study on the MAAPO algorithm

| Algorithm | MAAPO vs. | MS-APO | RFDB-APO | APO |
|---|---|---|---|---|
| +/=/– | – | 5/25/0 | 5/25/0 | 8/22/0 |

**Table 6** The simulation results of 13 algorithms on CEC2017

| Function | | MAAPO | APO | GWO | IGWO | PSO | PPSO | DE | ARO | FPA | GOA | MAQUATRE | SGA | AOO |
|---|---|---|---|---|---|---|---|---|---|---|---|---|---|---|
| F1 | Mean | 1.56E-05 | 1.72E-04(+) | 9.86E+06(+) | 1.32E+04(+) | 1.18E+02(+) | 3.25E+00(+) | 8.63E+02(+) | 3.05E+02(+) | 1.05E+03(+) | 2.81E+03(+) | **0.00E+00(−)** | 4.15E+08(+) | 2.65E+03(+) |
| | Std | 8.97E-05 | 6.70E-04 | 4.87E+07 | 4.99E+03 | 7.74E+02 | 1.54E+01 | 1.78E+03 | 4.61E+02 | 1.34E+03 | 2.95E+03 | **0.00E+00** | 6.52E+08 | 3.06E+03 |
| F2 | Mean | **0.00E+00** | **0.00E+00(=)** | 1.52E+06(+) | **0.00E+00(=)** | 4.41E+00(=) | **0.00E+00(=)** | 1.57E-01(+) | **0.00E+00(=)** | **0.00E+00(=)** | **0.00E+00(=)** | **0.00E+00(=)** | 1.35E+11(+) | 3.99E+01(+) |
| | Std | **0.00E+00** | **0.00E+00** | 7.17E+06 | **0.00E+00** | 2.50E+01 | **0.00E+00** | 4.18E-01 | **0.00E+00** | **0.00E+00** | **0.00E+00** | **0.00E+00** | 9.52E+11 | 3.86E+01 |
| F3 | Mean | 3.22E-05 | 3.66E-05(+) | 3.22E+02(+) | 3.54E-02(+) | **0.00E+00(−)** | **0.00E+00(−)** | 1.71E+03(+) | 4.71E-05(+) | **0.00E+00(−)** | 4.87E-08(−) | **0.00E+00(−)** | 6.11E+03(+) | 6.33E-05(+) |
| | Std | 1.43E-04 | 8.52E-05 | 7.62E+02 | 2.05E-02 | **0.00E+00** | **0.00E+00** | 5.77E+02 | 1.33E-04 | **0.00E+00** | 1.44E-07 | **0.00E+00** | 5.76E+03 | 6.71E-05 |
| F4 | Mean | 4.37E+00 | 4.58E+00(+) | 1.38E+01(+) | 2.10E+00(−) | 4.00E+00(−) | 1.16E-03(−) | 4.93E+00(+) | 4.02E+00(−) | 1.97E-01(−) | 7.70E-01(−) | **0.00E+00(−)** | 6.29E+01(+) | 4.41E+00(=) |
| | Std | 3.85E-01 | 5.07E-01 | 1.45E+01 | 3.76E-01 | 1.09E+01 | 8.27E-03 | 1.16E+00 | 9.48E-01 | 4.26E-02 | 5.11E-01 | **0.00E+00** | 5.76E+01 | 1.80E+00 |
| F5 | Mean | **1.22E+00** | 1.32E+00(=) | 1.24E+01(+) | 6.47E+00(+) | 3.04E+01(+) | 2.85E+01(+) | 6.39E+00(+) | 8.10E+00(+) | 3.37E+00(+) | 1.39E+01(+) | 5.09E+00(+) | 4.60E+01(+) | 1.82E+01(+) |
| | Std | **5.92E-01** | 6.90E-01 | 5.37E+00 | 4.77E+00 | 1.17E+01 | 9.46E+00 | 1.27E+00 | 3.15E+00 | 1.89E+00 | 6.76E+00 | 2.27E+00 | 1.54E+01 | 8.27E+00 |
| F6 | Mean | **0.00E+00** | **0.00E+00(=)** | 3.68E-01(+) | 4.47E-02(+) | 6.71E+00(+) | 5.81E+00(+) | **0.00E+00(=)** | 7.59E-06(+) | **0.00E+00(=)** | 2.78E-03(+) | **0.00E+00(=)** | 3.15E+01(+) | 6.89E-01(+) |
| | Std | **0.00E+00** | **0.00E+00** | 6.65E-01 | 1.65E-02 | 5.53E+00 | 5.46E+00 | **0.00E+00** | 2.17E-05 | **0.00E+00** | 7.81E-03 | **0.00E+00** | 1.25E+01 | 9.91E-01 |
| F7 | Mean | **1.14E+01** | 1.15E+01(=) | 2.44E+01(+) | 2.28E+01(+) | 2.80E+01(+) | 2.65E+01(+) | 1.75E+01(+) | 2.04E+01(+) | 2.10E+01(+) | 1.98E+01(+) | 1.53E+01(+) | 7.62E+01(+) | 3.07E+01(+) |
| | Std | **4.67E-01** | 6.23E-01 | 7.86E+00 | 8.22E+00 | 9.29E+00 | 7.24E+00 | 2.11E+00 | 3.76E+00 | 3.85E+00 | 4.49E+00 | 2.65E+00 | 3.50E+01 | 7.96E+00 |
| F8 | Mean | **1.16E+00** | 1.52E+00(+) | 1.12E+01(+) | 6.27E+00(+) | 1.92E+01(+) | 1.55E+01(+) | 6.48E+00(+) | 9.19E+00(+) | 2.68E+00(+) | 1.22E+01(+) | 5.34E+00(+) | 3.61E+01(+) | 1.61E+01(+) |
| | Std | **5.98E-01** | 7.91E-01 | 5.31E+00 | 3.75E+00 | 7.94E+00 | 5.84E+00 | 1.38E+00 | 3.65E+00 | 1.79E+00 | 6.39E+00 | 2.05E+00 | 1.65E+01 | 6.13E+00 |
| F9 | Mean | **0.00E+00** | **0.00E+00(=)** | 4.41E+00(+) | 6.96E-04(+) | 2.93E-01(+) | 1.53E-01(+) | **0.00E+00(=)** | 3.64E-04(=) | **0.00E+00(=)** | 9.26E-02(=) | **0.00E+00(=)** | 4.56E+02(+) | 7.63E-01(+) |
| | Std | **0.00E+00** | **0.00E+00** | 1.18E+01 | 4.93E-04 | 5.47E-01 | 2.24E-01 | **0.00E+00** | 2.59E-03 | **0.00E+00** | 6.36E-01 | **0.00E+00** | 2.82E+02 | 3.71E+00 |
| F10 | Mean | 1.62E+02 | **1.57E+02(=)** | 4.61E+02(+) | 4.30E+02(+) | 9.09E+02(+) | 8.48E+02(+) | 2.89E+02(+) | 2.70E+02(+) | 7.12E+02(+) | 6.54E+02(+) | 1.67E+02(=) | 1.30E+03(+) | 6.27E+02(+) |
| | Std | 1.03E+02 | **8.70E+01** | 3.02E+02 | 3.12E+02 | 2.69E+02 | 2.93E+02 | 1.03E+02 | 1.51E+02 | 2.46E+02 | 2.49E+02 | 1.38E+02 | 3.61E+02 | 2.39E+02 |
| F11 | Mean | 2.16E+00 | 2.11E+00(=) | 2.31E+01(+) | 2.95E+00(+) | 3.48E+01(+) | 2.40E+01(+) | 2.46E+00(=) | 3.52E+00(+) | 2.98E+00(+) | 7.10E+00(+) | **8.09E-01(−)** | 1.96E+02(+) | 2.46E+01(+) |
| | Std | **5.18E-01** | 5.67E-01 | 1.26E+01 | 2.13E+00 | 1.73E+01 | 1.61E+01 | 1.06E+00 | 2.73E+00 | 1.18E+00 | 3.85E+00 | 1.13E+00 | 1.63E+02 | 1.16E+01 |
| F12 | Mean | 1.22E+02 | 2.30E+02(+) | 5.74E+05(+) | 2.25E+04(+) | 4.23E+03(+) | 3.63E+03(+) | 7.60E+04(+) | 5.05E+03(+) | 1.55E+04(+) | 1.65E+04(+) | **1.16E+02(=)** | 4.46E+07(+) | 4.51E+05(+) |
| | Std | 2.09E+02 | 3.58E+02 | 7.84E+05 | 1.71E+04 | 4.25E+03 | 7.37E+03 | 4.02E+04 | 5.39E+03 | 1.41E+04 | 1.50E+04 | **9.98E+01** | 1.72E+08 | 4.62E+05 |
| F13 | Mean | **5.63E+00** | 6.20E+00(+) | 8.56E+03(+) | 8.48E+02(+) | 8.41E+02(+) | 3.41E+02(+) | 1.66E+02(+) | 6.66E+00(=) | 1.48E+03(+) | 1.41E+03(+) | 5.72E+00(=) | 1.76E+04(+) | 1.08E+04(+) |
| | Std | 1.49E+00 | **1.47E+00** | 5.29E+03 | 1.13E+03 | 9.42E+02 | 2.48E+02 | 1.45E+02 | 5.88E+00 | 1.38E+03 | 1.58E+03 | 3.05E+00 | 2.15E+04 | 8.64E+03 |
| F14 | Mean | **1.03E-01** | 2.03E-01(=) | 9.23E+02(+) | 5.10E+01(+) | 4.96E+01(+) | 5.17E+01(+) | 9.40E-01(+) | 3.76E+00(+) | 5.64E+01(+) | 4.03E+01(+) | 4.25E+00(+) | 5.50E+03(+) | 8.33E+01(+) |
| | Std | **1.68E-01** | 3.01E-01 | 1.54E+03 | 1.03E+01 | 2.21E+01 | 1.90E+01 | 5.65E-01 | 2.72E+00 | 1.13E+01 | 1.53E+01 | 7.63E+00 | 7.58E+03 | 3.44E+01 |
| F15 | Mean | 4.95E-01 | 5.65E-01(=) | 1.43E+03(+) | 3.24E+01(+) | 6.84E+01(+) | 5.74E+01(+) | 1.15E+00(+) | 2.10E+00(+) | 1.19E+02(+) | 3.59E+01(+) | **3.56E-01(−)** | 1.60E+04(+) | 1.57E+02(+) |
| | Std | 2.95E-01 | **2.61E-01** | 1.55E+03 | 1.58E+01 | 5.05E+01 | 4.35E+01 | 5.07E-01 | 1.97E+00 | 6.21E+01 | 2.99E+01 | 4.98E-01 | 2.57E+04 | 9.60E+01 |
| F16 | Mean | **8.63E-01** | 9.13E-01(=) | 6.46E+01(+) | 9.25E+00(+) | 1.86E+02(+) | 1.52E+02(+) | 1.84E+00(+) | 3.96E+01(=) | 2.89E+00(+) | 5.07E+01(+) | 8.94E-01(−) | 2.99E+02(+) | 6.56E+01(+) |
| | Std | 3.64E-01 | **2.69E-01** | 5.76E+01 | 2.43E+01 | 1.20E+02 | 9.85E+01 | 2.25E+00 | 6.68E+01 | 1.39E+00 | 8.44E+01 | 2.18E+00 | 1.55E+02 | 8.56E+01 |

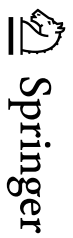

**Table 6** (continued)

| Function | | MAAPO | APO | GWO | IGWO | PSO | PPSO | DE | ARO | FPA | GOA | MAQUATRE | SGA | AOO |
|---|---|---|---|---|---|---|---|---|---|---|---|---|---|---|
| F17 | Mean | 1.50E+00 | 1.41E+00(=) | 5.31E+01(+) | 3.49E+01(+) | 5.48E+01(+) | 4.67E+01(+) | **1.05E+00(−)** | 1.24E+01(+) | 2.82E+01(+) | 3.18E+01(+) | 2.23E+00(=) | 1.15E+02(+) | 4.50E+01(+) |
| | Std | 7.74E-01 | **5.97E-01** | 3.01E+01 | 9.47E+00 | 2.93E+01 | 1.18E+01 | 3.91E+00 | 1.50E+01 | 9.98E+00 | 2.15E+01 | 4.06E+00 | 7.00E+01 | 1.60E+01 |
| F18 | Mean | **4.34E-01** | 4.68E-01(=) | 2.50E+04(+) | 5.09E+03(+) | 3.55E+03(+) | 9.45E+02(+) | 7.01E+00(+) | 1.14E+00(+) | 8.28E+03(+) | 3.40E+03(+) | 7.32E+00(+) | 1.53E+04(+) | 2.28E+04(+) |
| | Std | 4.93E-01 | **4.63E-01** | 1.47E+04 | 4.95E+03 | 9.53E+03 | 4.68E+03 | 8.00E+00 | 1.03E+00 | 6.27E+03 | 4.09E+03 | 9.60E+00 | 1.54E+04 | 1.32E+04 |
| F19 | Mean | 1.08E-01 | 1.02E-01(=) | 4.40E+03(+) | 2.42E+01(+) | 2.40E+01(+) | 2.57E+01(+) | **5.88E-02(−)** | 7.39E-01(+) | 9.57E+01(+) | 2.21E+01(+) | 1.51E-01(−) | 3.27E+05(+) | 6.76E+01(+) |
| | Std | 5.76E-02 | **5.01E-02** | 5.54E+03 | 7.89E+00 | 1.96E+01 | 2.04E+01 | 2.23E-01 | 6.43E-01 | 1.20E+02 | 1.57E+01 | 3.78E-01 | 1.46E+06 | 1.01E+02 |
| F20 | Mean | 3.36E-01 | **2.32E-01(=)** | 5.28E+01(+) | 2.45E+01(+) | 1.01E+02(+) | 7.75E+01(+) | 2.82E-01(=) | 4.15E+00(+) | 8.32E+00(+) | 2.97E+01(+) | 2.47E+00(+) | 1.68E+02(+) | 5.71E+01(+) |
| | Std | 2.85E-01 | 2.59E-01 | 3.55E+01 | 8.13E+00 | 5.78E+01 | 3.97E+01 | **2.09E-01** | 5.90E+00 | 9.34E+00 | 1.95E+01 | 5.33E+00 | 8.38E+01 | 4.41E+01 |
| F21 | Mean | **1.16E+02** | 1.29E+02(=) | 1.88E+02(+) | 1.91E+02(+) | 1.84E+02(+) | 1.39E+02(=) | 1.27E+02(+) | 1.35E+02(=) | 1.73E+02(+) | 1.75E+02(+) | 1.63E+02(+) | 1.78E+02(+) | 1.22E+02(=) |
| | Std | **3.00E+01** | 3.93E+01 | 5.03E+01 | 3.98E+01 | 6.32E+01 | 5.82E+01 | 4.53E+01 | 5.15E+01 | 4.94E+01 | 5.48E+01 | 5.33E+01 | 6.72E+01 | 4.39E+01 |
| F22 | Mean | 9.89E+01 | 9.89E+01(+) | 1.07E+02(+) | 1.05E+02(+) | 9.57E+01(+) | 9.74E+01(+) | **8.58E+01(+)** | 9.50E+01(+) | 1.00E+02(+) | 9.65E+01(+) | 9.37E+01(+) | 1.58E+02(+) | 9.76E+01(+) |
| | Std | 7.81E+00 | 7.63E+00 | 6.74E+00 | 9.83E-01 | 2.26E+01 | 1.76E+01 | 3.26E+01 | 2.13E+01 | **1.32E-01** | 1.85E+01 | 2.28E+01 | 5.06E+01 | 2.11E+01 |
| F23 | Mean | **3.04E+02** | **3.04E+02(=)** | 3.13E+02(+) | 3.10E+02(+) | 3.59E+02(+) | 3.40E+02(+) | 3.09E+02(+) | 3.11E+02(+) | 3.07E+02(+) | 3.17E+02(+) | 3.05E+02(+) | 3.50E+02(+) | 3.22E+02(+) |
| | Std | 1.39E+00 | **1.16E+00** | 7.70E+00 | 5.84E+00 | 2.34E+01 | 5.15E+01 | 1.61E+00 | 5.17E+00 | 2.61E+00 | 8.94E+00 | 1.95E+00 | 1.93E+01 | 7.04E+00 |
| F24 | Mean | 2.96E+02 | 2.86E+02(=) | 3.43E+02(+) | 3.34E+02(+) | 3.29E+02(+) | **2.11E+02(−)** | 2.68E+02(=) | 2.43E+02(=) | 3.36E+02(+) | 3.31E+02(+) | 2.84E+02(=) | 3.72E+02(+) | 3.23E+02(+) |
| | Std | 7.41E+01 | 9.39E+01 | 1.18E+01 | 3.40E+01 | 1.09E+02 | 1.30E+02 | 9.66E+01 | 1.18E+02 | **3.13E+00** | 6.84E+01 | 9.76E+01 | 5.28E+01 | 7.21E+01 |
| F25 | Mean | 4.18E+02 | 4.15E+02(=) | 4.33E+02(+) | **3.99E+02(−)** | 4.16E+02(=) | 4.00E+02(−) | 4.00E+02(−) | 4.14E+02(=) | 4.24E+02(=) | 4.28E+02(+) | 4.17E+02(=) | 4.97E+02(+) | 4.16E+02(=) |
| | Std | 2.31E+01 | 2.26E+01 | 1.74E+01 | **4.66E+00** | 5.06E+01 | 4.59E+01 | 6.68E+00 | 2.16E+01 | 2.32E+01 | 2.64E+01 | 2.32E+01 | 7.15E+01 | 2.25E+01 |
| F26 | Mean | 3.09E+02 | 3.01E+02(=) | 3.41E+02(+) | 3.30E+02(=) | 3.37E+02(+) | **2.28E+02(−)** | 2.86E+02(=) | 3.00E+02(+) | 3.17E+02(=) | 4.01E+02(+) | 2.98E+02(−) | 7.09E+02(+) | 2.98E+02(=) |
| | Std | 2.17E+01 | 4.59E+01 | 1.35E+02 | 1.81E+02 | 1.01E+02 | 1.04E+02 | 6.28E+01 | 1.70E+01 | 2.91E+01 | 2.05E+02 | **1.40E+01** | 3.50E+02 | 2.89E+01 |
| F27 | Mean | **3.89E+02** | 3.90E+02(+) | 3.93E+02(+) | **3.89E+02(=)** | 4.35E+02(+) | 4.08E+02(+) | **3.89E+02(=)** | 3.94E+02(+) | 3.90E+02(+) | 3.95E+02(+) | 3.90E+02(=) | 4.36E+02(+) | 3.92E+02(+) |
| | Std | 3.78E-01 | **3.20E-01** | 4.50E+00 | 3.89E-01 | 3.82E+01 | 1.99E+01 | 7.30E-01 | 2.56E+00 | 5.19E-01 | 1.23E+01 | 1.72E+00 | 3.34E+01 | 2.19E+00 |
| F28 | Mean | 4.52E+02 | 4.26E+02(=) | 5.27E+02(+) | 4.76E+02(=) | 4.71E+02(=) | 3.61E+02(−) | 3.61E+02(−) | **3.05E+02(−)** | 4.06E+02(=) | 5.41E+02(+) | 3.25E+02(−) | 5.63E+02(+) | 4.21E+02(=) |
| | Std | 1.20E+02 | 1.08E+02 | 1.06E+02 | 1.55E+02 | 1.36E+02 | 1.12E+02 | 9.44E+01 | 8.91E+01 | 1.11E+02 | 2.10E+02 | **6.58E+01** | 1.50E+02 | 1.16E+02 |
| F29 | Mean | **2.39E+02** | 2.40E+02(=) | 2.79E+02(+) | 2.49E+02(+) | 3.19E+02(+) | 2.97E+02(+) | 2.51E+02(+) | 2.54E+02(+) | 2.60E+02(+) | 2.70E+02(+) | 2.43E+02(+) | 4.59E+02(+) | 2.73E+02(+) |
| | Std | **3.29E+00** | 5.31E+00 | 3.92E+01 | 9.79E+00 | 3.92E+01 | 3.14E+01 | 6.75E+00 | 1.63E+01 | 1.25E+01 | 4.20E+01 | 9.32E+00 | 1.01E+02 | 2.54E+01 |
| F30 | Mean | 1.50E+04 | 1.90E+03(=) | 3.88E+05(+) | 8.03E+04(+) | 2.44E+05(+) | **1.70E+03(=)** | 6.00E+03(+) | 2.03E+03(+) | 1.62E+05(+) | 1.38E+05(+) | 2.41E+04(−) | 2.64E+06(+) | 4.39E+04(+) |
| | Std | 9.06E+04 | 3.62E+03 | 5.65E+05 | 2.36E+05 | 5.03E+05 | 2.23E+03 | 4.69E+03 | **1.67E+03** | 2.79E+05 | 3.24E+05 | 1.26E+05 | 3.94E+06 | 1.58E+05 |
| Average ranking | | 3.35 | 3.55 | 10.60 | 7.32 | 9.35 | 7.10 | 4.43 | 5.25 | 6.90 | 7.95 | 3.37 | 12.80 | 9.03 |
| +/=/− | | – | 8/22/0 | 30/0/0 | 24/4/2 | 25/3/2 | 21/3/6 | 19/7/4 | 21/7/2 | 22/6/2 | 26/2/2 | 10/10/10 | 30/0/0 | 25/5/0 |

The best results are shown in bold

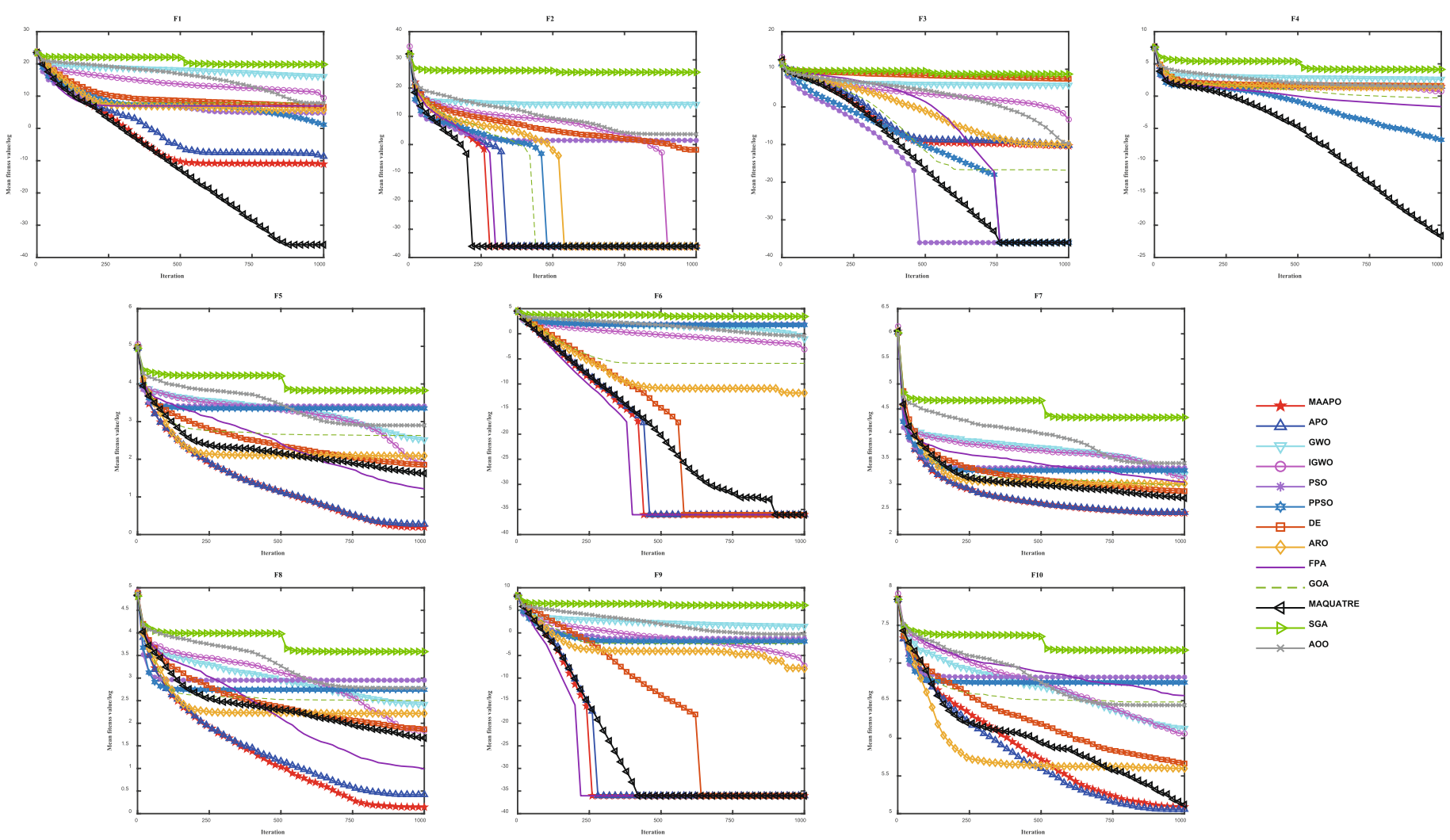

**Fig. 4** The convergence curves of 13 algorithms on F1–F10

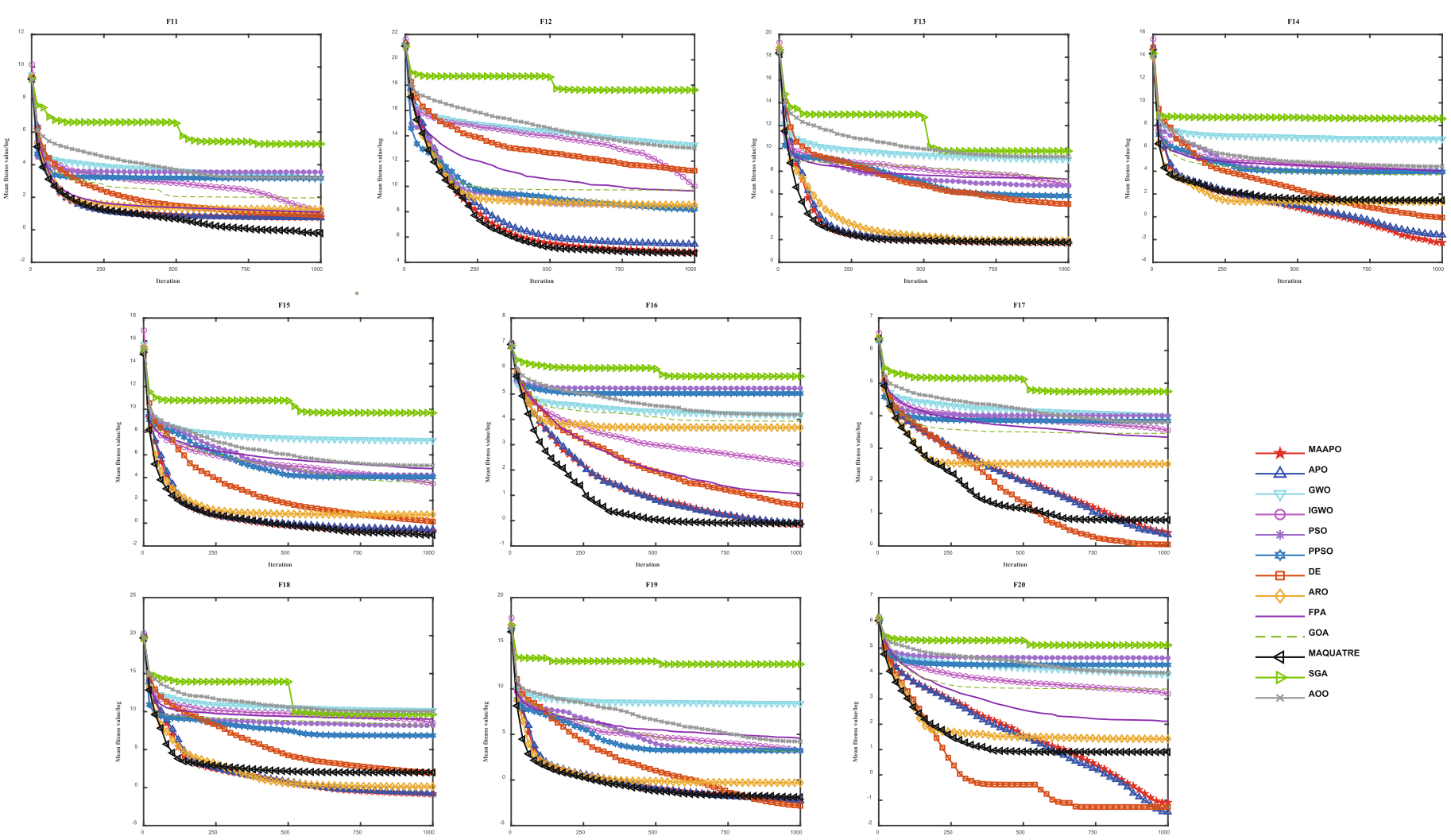

**Fig. 5** The convergence curves of 13 algorithms on F11–F20

9.03, PSO with 9.35, GWO with 10.60, and SGA with 12.80. Figs. 4, 5, 6 plot the convergence curves based on mean fitness values. MAAPO shows superior convergence speed and accuracy on most test functions, whereas the other algorithms tend to converge prematurely. Table 7 records the running time of the algorithms. The time of MAPPO is improved compared with the original APO, but it is acceptable considering the better search performance. As a result, the test results on CEC2017 confirm the excellent performance of MAAPO.

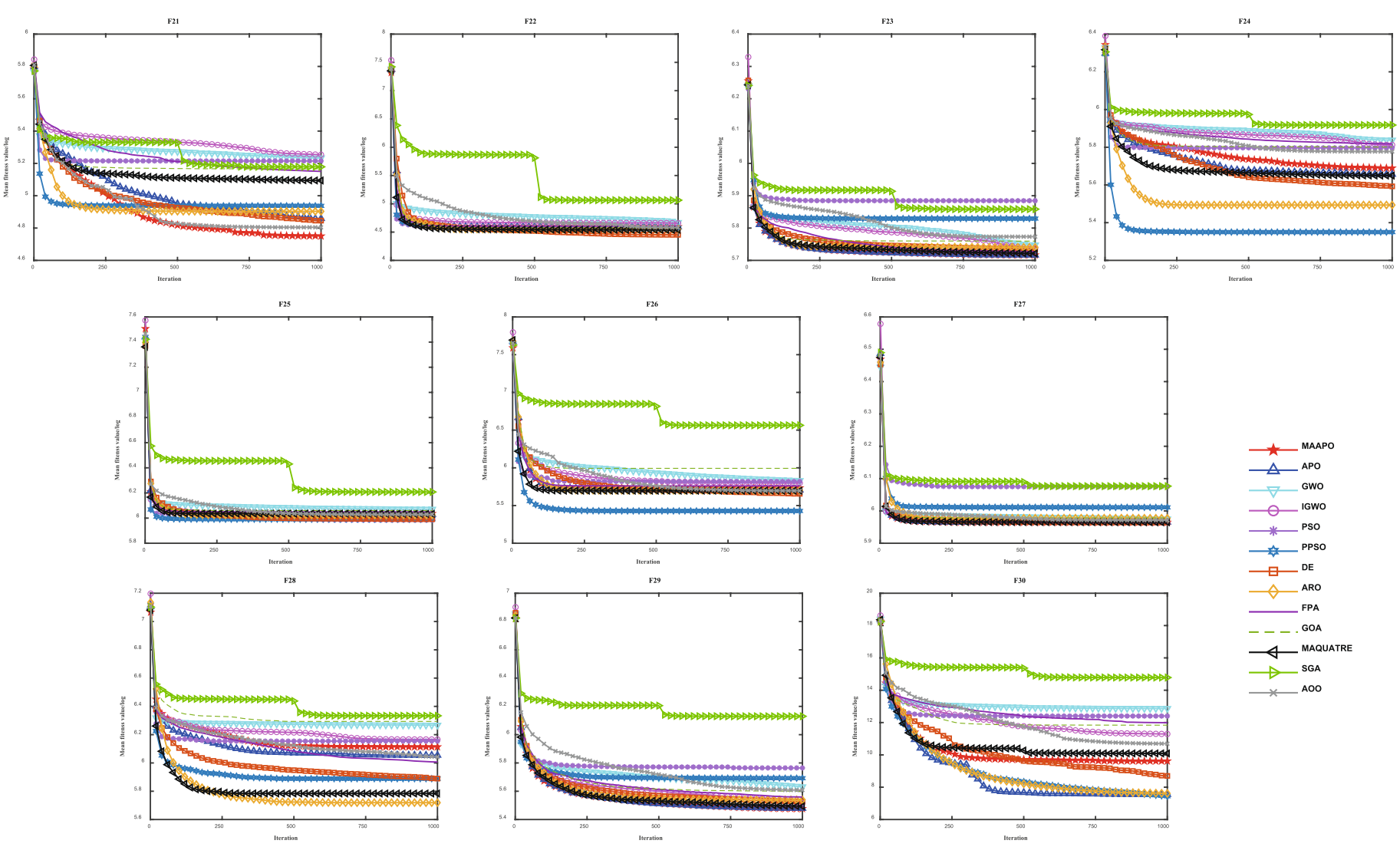

**Fig. 6** The convergence curves of 13 algorithms on F21–F30

## 4.6 Analysis of the MAAPO algorithm

The MAAPO algorithm is highly competitive, not only benefiting from the excellent APO algorithm but also from the improved strategies introduced in this study. First, the RFDB selection mechanism updates the random selection of $X_j$ in autotrophic foraging. This selection is a probability-based method that takes into account greediness and randomness while maintaining the exploratory character of the autotrophic foraging model. This allows potential solutions to be selected as reference points to guide the population. Moreover, the $M_f$ of the APO algorithm is a forage mapping vector that is affected by the population size. When the population size decreases, the crossover ratio of dimensions increases, thus improving the algorithm's exploration. The membrane system introduces a separating-merging principle. The separation operation will generate multiple membranes to randomly divide the population, making the individuals within each elementary membrane smaller than the one membrane and thus promoting diversity. The maximum number of separation membranes is set to 4. This dynamic separation strategy is more random than the static number of membranes, and the results are more competitive. The merging operation also promotes information exchange among the population, regardless of whether the next update operation is based on one membrane or multiple membranes. As the iteration proceeds, the population will gradually converge. This makes the parameter $mVOL$ gradually decrease, and the algorithm transforms from the membrane separating-merging to one membrane operation, supporting the balance between exploration and exploitation. Experiments have confirmed that the combined strategies are effective in improving performance.

**Table 7** Average runtime (s) of each algorithm on the CEC2017 test suite

| | MAAPO | APO | GWO | IGWO | PSO | PPSO | DE | ARO | FPA | GOA | MAQUATRE | SGA | AOO |
|---|---|---|---|---|---|---|---|---|---|---|---|---|---|
| Runtime | 1.38 | 0.81 | 0.79 | 1.20 | 0.18 | 2.93 | 0.84 | 1.07 | 1.29 | 0.99 | 0.63 | 0.23 | 1.14 |

## 5 Multilevel thresholding image segmentation

This section applies the proposed MAAPO algorithm to color image segmentation by optimizing multiple thresholds using the Otsu and Kapur entropy methods. Segmentation quality is evaluated through PSNR, SSIM, and FSIM.

### 5.1 Fitness functions

Otsu and Kapur entropy methods both maximize fitness functions to obtain multiple thresholds, and then each class can be represented by averaging the pixel values. The detailed fitness function values are introduced as follows.

- Otsu method

The Otsu method (Pare et al. 2016; Ma and Yue 2022; Huang et al. 2021; Jia et al. 2024) maximizes the variance of distinct classes to search for the thresholds. For multilevel thresholding image segmentation, the fitness function is as follows:

$$\{t_1^*, t_2^*, \ldots, t_{n-1}^*\} = \arg\max f(t_1, \ldots, t_{n-1}), \quad \text{s.t. } 0 < t_1 < t_2 < \cdots < t_{n-1} < L-1 \tag{22}$$

$$f = \sigma_0 + \sigma_1 + \cdots + \sigma_{n-1} \tag{23}$$

$$\sigma_0 = \omega_0 \cdot (\mu_0 - \mu_T)^2, \quad \sigma_1 = \omega_1 \cdot (\mu_1 - \mu_T)^2, \quad \sigma_{n-1} = \omega_{n-1} \cdot (\mu_{n-1} - \mu_T)^2 \tag{24}$$

$$\omega_0 = \sum_{i=0}^{t_1-1} p(i), \quad \omega_1 = \sum_{i=t_1}^{t_2-1} p(i), \quad \omega_{n-1} = \sum_{i=t_{n-1}}^{L-1} p(i) \tag{25}$$

$$\mu_0 = \frac{\sum_{i=0}^{t_1-1} i \cdot p(i)}{\omega_0}, \quad \mu_1 = \frac{\sum_{i=t_1}^{t_2-1} i \cdot p(i)}{\omega_1}, \quad \mu_{n-1} = \frac{\sum_{i=t_{n-1}}^{L-1} i \cdot p(i)}{\omega_{n-1}} \tag{26}$$

$$\mu_T = \sum_{i=t_0}^{L-1} i \cdot p(i), \quad p(i) = \frac{h_i}{N} \tag{27}$$

where:

- $L$ represents the intensity level of a grayscale image or each color image band.
- $p(i)$ denotes the $i$ pixel value's probability distribution.
- $h_i$ is the histogram number of the $i$ pixel value.
- $N$ is the total histogram number of pixel values.

- $n-1$ is the number of thresholds for segmenting the image.Kapur entropy method

Entropy measures the uncertainty or randomness of information distribution. In image segmentation, entropy can quantify the compactness and separability among classes. The Kapur entropy method (Kapur et al. 1985; Abdel-Basset et al. 2022; Upadhyay and Chhabra 2020) calculates the probability distribution of the image histogram to determine the thresholds. Its fitness function can be expressed as Eq. (22).

$$f = K_0 + K_1 + \cdots + K_{n-1} \tag{28}$$

$$K_0 = -\sum_{i=0}^{t_1-1} \frac{p(i)}{\omega_0} \ln \frac{p(i)}{\omega_0}, \quad K_1 = -\sum_{t_1}^{t_2-1} \frac{p(i)}{\omega_1} \ln \frac{p(i)}{\omega_1}, \quad K_{n-1} = -\sum_{t_{n-1}}^{L-1} \frac{p(i)}{\omega_{n-1}} \ln \frac{p(i)}{\omega_{n-1}} \tag{29}$$

where:

- $\omega_0$, $\omega_1$, and $\omega_{n-1}$ refer to Eq. (25).
- $p(i)$ refer to Eq. (27).

### 5.2 Evaluation metrics

Three quantitative evaluation metrics are introduced to assess segmentation quality. Higher metric values reflect better results. Their details are provided below.

- PSNR

PSNR is a metric that measures the difference between the original signal and the processed signal. In image segmentation, it can evaluate image quality by comparing the peak signal-to-noise ratio between the original and the segmented images. A higher PSNR value indicates that there is less distortion and the segmented image is quite similar to the original image.

$$\text{PSNR} = 20 \cdot \log_{10} \left( \frac{255}{\sqrt{\text{MSE}}} \right) \tag{30}$$

$$\text{MSE} = \frac{1}{m \cdot n} \sum_{i=0}^{m-1} \sum_{j=0}^{n-1} [I(i,j) - I'(i,j)]^2 \tag{31}$$

where:

- MSE is the mean square error between the original and segmented images.
- $m \cdot n$ is the image size.
- $I$ means the original image.

- $I'$ denotes the segmented image.SSIM

SSIM is a measure of the similarity between two images, which considers more structural information in images. It more closely matches how the human eye judges the quality of an image. Three factors are compared for similarity: luminance, contrast, and structure.

$$\mathrm{SSIM}(I, I') = \frac{(2\mu_I \mu_{I'} + C_1)(\sigma_{II'} + C_2)}{(\mu_I^2 + \mu_{I'}^2 + C_1)(\sigma_I^2 + \sigma_{I'}^2 + C_2)} \tag{32}$$

where:

- $\mu_I$ and $\mu_{I'}$ are the mean values of the images $I$ and $I'$, respectively.
- $\sigma_I^2$ and $\sigma_{I'}^2$ are the variances of the images $I$ and $I'$, respectively.
- $\sigma_{II'}$ is the covariance of the images $I$ and $I'$.

- $C_1$ is set to 6.5025, and $C_2$ is set to 8.5225.FSIM

FSIM can effectively evaluate image quality by measuring the similarity between two images. It is particularly concerned with the image feature information including phase congruency and gradient magnitude.

$$\mathrm{FSIM}(I, I') = \frac{\sum_{x \in \Omega} S_L(x) \cdot PC_m(x)}{\sum_{x \in \Omega} PC_m(x)} \tag{33}$$

$$PC_m = \max(PC_1 + PC_2) \tag{34}$$

$$S_L(x) = S_{PC}(x) \cdot S_G(x) \tag{35}$$

$$S_{PC}(x) = \frac{2PC_1(x) \cdot PC_2(x) + T_1}{PC_1^2(x) + PC_2^2(x) + T_1} \tag{36}$$

$$S_G(x) = \frac{2G_1(x) \cdot G_2(x) + T_2}{G_1^2(x) + G_2^2(x) + T_2} \tag{37}$$

where:

- $\Omega$ denotes the image's entire spatial domain.
- $x$ is the pixel point.
- $PC_1$ and $PC_2$ are the phase consistency of the images $I$ and $I'$, respectively.
- $G_1$ and $G_2$ are the gradients of the images $I$ and $I'$, respectively.
- $T_1$ is set to 0.85, and $T_2$ is set to 160.

### 5.3 Performance evaluation of MAAPO on image segmentation

This subsection evaluates the performance of MAAPO and compares it with APO, GWO, PSO, DE, MAQUATRE, SGA, and AOO. The experiment tested three images: the parrot image has a resolution of 768 $\times$ 512 pixels, while the owl and horse images each have a resolution of 481$\times$321 pixels. Fig. 7 displays the original color images and their histograms. Each algorithm is executed 20 times, while the remaining parameters remain identical to the experimental setup in 4.1. Experiments with 4, 5, 6, and 7 thresholds were conducted for each image, and the optimal result for each algorithm was recorded. Tables 8, 9, 10are the results of the test images based on the Otsu and Kapur entropy methods. Each table lists the thresholds on the RGB bands and the values of the three evaluation metrics. The best results are shown in bold. The most tested algorithms performed similarly in the case of 4 thresholds, whether based on Otsu or Kapur entropy. This is because the algorithms are comparable due to the low dimensionality of the problem. As the dimensionality increases, the proposed MAAPO algorithm performs outstandingly and beats its comparison algorithms in most instances.

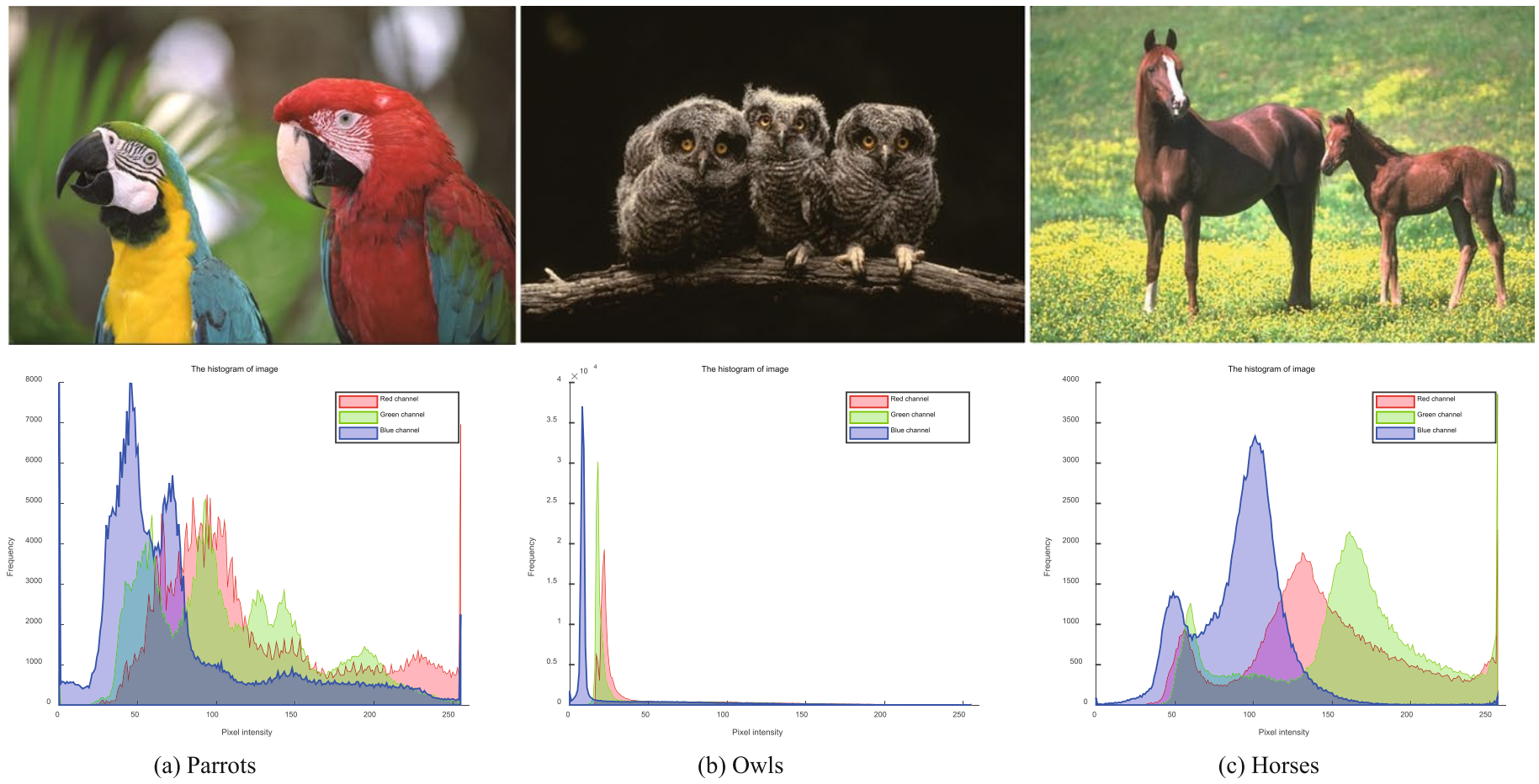

(a) Parrots (b) Owls (c) Horses

**Fig. 7** Test color images and their histograms

Figures 8 and 9 are the segmented images of the MAAPO algorithm at different thresholds. The image segmentation process is as follows: MAAPO is used to search the thresholds of each RGB band, Otsu and Kapur entropy are adopted as fitness functions, and the outcomes of the three band segmentation are then cascaded. The Friedman test method was used to calculate the average rank of each algorithm on the three images. Tables 11 and 12 provide an overview of the outcomes. According to Table 11, the MAAPO algorithm outperforms others in SSIM under the Otsu method, with an average rank of 3.83. The best ranking in the PSNR metric is PSO, with an average ranking of 3.17. The best ranking in the FSIM metric is DE, with an average ranking of 3.58. MAAPO provides outcomes that are superior to the original APO algorithm, although not achieve the best performance in these two metrics. In Table 12, the proposed MAAPO algorithm achieves the best overall performance when applied with the Kapur entropy method. The average ranks of PSNR, SSIM, and FSIM for the algorithm are 3.17, 3.75, and 3.42, respectively. The experimental results demonstrate that the MAAPO algorithm performs competitively among the evaluated optimizers and effectively solves the image segmentation problem.

## 6 Conclusion

This study proposes a novel variant of the artificial protozoa optimizer (APO), named MAAPO. The proposed algorithm enhances optimization performance through two key improvements. First, a membrane system framework is incorporated to introduce a parallel distributed paradigm to improve population diversity and search dynamics. To support

**Table 8** The best thresholds of algorithms in the parrot image

| Method | Algorithm | Thresholds | Red band | Green band | Blue band | PSNR | SSIM | FSIM |
|---|---|---|---|---|---|---|---|---|
| Otsu | MAAPO | 4 | 82 120 165 213 | 74 115 159 201 | 36 64 112 179 | **26.50468** | 0.82025 | **0.85764** |
| | | 5 | 73 99 130 171 216 | 72 106 135 169 208 | 27 57 89 133 188 | **28.04244** | **0.84965** | 0.88528 |
| | | 6 | 70 93 116 145 180 219 | 55 80 108 136 169 208 | 26 55 85 123 165 208 | 29.21866 | **0.86934** | 0.91443 |
| | | 7 | 69 90 109 134 167 201 231 | 52 72 93 115 138 170 209 | 20 43 62 87 124 166 209 | 30.32407 | 0.89175 | 0.93186 |
| | APO | 4 | 82 120 165 213 | 74 115 159 201 | 36 64 112 179 | **26.50468** | 0.82025 | **0.85764** |
| | | 5 | 73 98 129 170 215 | 72 106 135 169 208 | 27 57 89 134 189 | 28.04154 | 0.84958 | 0.88542 |
| | | 6 | 71 94 117 145 180 220 | 57 81 108 135 169 208 | 27 56 84 122 165 208 | 29.22007 | 0.86816 | **0.91577** |
| | | 7 | 69 91 112 136 168 202 231 | 52 73 94 115 139 170 210 | 20 43 62 89 128 169 211 | 30.32227 | 0.89135 | 0.93003 |
| | GWO | 4 | 82 120 165 213 | 74 115 159 201 | 36 64 112 179 | **26.50468** | 0.82025 | **0.85764** |
| | | 5 | 73 99 130 171 216 | 72 106 135 169 208 | 27 57 89 133 188 | **28.04244** | **0.84965** | 0.88528 |
| | | 6 | 70 93 117 145 180 219 | 56 80 108 136 169 208 | 27 56 85 123 165 208 | **29.22260** | 0.86863 | 0.91491 |
| | | 7 | 69 91 111 136 168 201 231 | 53 74 95 116 140 171 209 | 20 43 62 88 125 167 209 | **30.33329** | 0.89175 | 0.93270 |
| | PSO | 4 | 82 120 165 213 | 74 115 159 201 | 36 64 112 179 | **26.50468** | 0.82025 | **0.85764** |
| | | 5 | 73 99 130 171 216 | 72 106 135 169 208 | 27 57 89 133 188 | **28.04244** | **0.84965** | 0.88528 |
| | | 6 | 70 93 117 145 180 219 | 56 80 108 136 169 208 | 27 56 85 123 165 208 | **29.22260** | 0.86863 | 0.91491 |
| | | 7 | 69 91 111 136 168 201 231 | 53 74 95 116 140 171 209 | 20 43 62 88 125 167 209 | **30.33329** | 0.89175 | 0.93270 |
| | DE | 4 | 82 121 166 213 | 74 115 159 202 | 36 63 112 180 | 26.50374 | **0.82031** | 0.85747 |
| | | 5 | 73 98 128 170 215 | 72 106 135 168 209 | 29 56 88 133 189 | 28.03484 | 0.84879 | **0.88572** |

**Table 8** (continued)

| Method | Algorithm | Thresholds | Red band | Green band | Blue band | PSNR | SSIM | FSIM |
|---|---|---|---|---|---|---|---|---|
| | | 6 | 73 97 121 150 186 221 | 55 78 107 136 171 211 | 28 56 85 124 167 209 | 29.19714 | 0.86676 | 0.91488 |
| | | 7 | 70 91 114 137 167 199 232 | 55 75 96 117 139 171 210 | 19 42 62 88 126 165 208 | 30.30106 | **0.89218** | **0.93344** |
| | MAQUATRE | 4 | 82 120 165 213 | 74 115 159 201 | 36 64 112 179 | **26.50468** | 0.82025 | **0.85764** |
| | | 5 | 73 99 130 171 216 | 72 106 135 169 208 | 27 57 89 133 188 | **28.04244** | **0.84965** | 0.88528 |
| | | 6 | 70 93 117 145 180 219 | 56 80 108 136 169 208 | 27 56 85 123 165 208 | **29.22260** | 0.86863 | 0.91491 |
| | | 7 | 69 91 111 136 168 201 231 | 53 74 95 116 140 171 209 | 20 43 62 88 125 167 209 | **30.33329** | 0.89175 | 0.93270 |
| | SGA | 4 | 82 120 165 213 | 74 115 159 201 | 36 64 112 179 | **26.50468** | 0.82025 | **0.85764** |
| | | 5 | 73 99 130 171 216 | 72 106 135 169 208 | 27 57 89 133 188 | **28.04244** | **0.84965** | 0.88528 |
| | | 6 | 70 93 117 145 180 219 | 56 80 108 136 169 208 | 27 56 85 123 165 208 | **29.22260** | 0.86863 | 0.91491 |
| | | 7 | 70 91 112 137 169 204 233 | 54 75 97 118 140 171 209 | 20 43 62 88 125 167 209 | 30.32585 | 0.89162 | 0.93209 |
| | AOO | 4 | 82 120 165 213 | 74 115 159 201 | 36 64 112 179 | **26.50468** | 0.82025 | **0.85764** |
| | | 5 | 73 99 130 171 216 | 72 106 135 169 208 | 27 57 89 133 188 | **28.04244** | **0.84965** | 0.88528 |
| | | 6 | 70 93 117 145 180 219 | 56 80 108 136 169 208 | 27 56 85 123 165 208 | **29.22260** | 0.86863 | 0.91491 |
| | | 7 | 69 91 111 136 168 201 231 | 53 74 95 116 140 171 209 | 20 43 62 88 125 167 209 | **30.33329** | 0.89175 | 0.93270 |
| Kapur | MAAPO | 4 | 76 117 160 206 | 74 115 158 207 | 82 121 163 204 | **25.27021** | 0.80767 | 0.85459 |
| | | 5 | 55 89 124 166 210 | 72 111 151 183 214 | 51 84 122 163 204 | **27.35636** | 0.83860 | 0.89495 |
| | | 6 | 55 88 120 154 188 220 | 66 100 129 157 184 214 | 45 82 112 145 178 212 | **28.08316** | **0.84955** | **0.90962** |

**Table 8** (continued)

| Method | Algorithm | Thresholds | Red band | Green band | Blue band | PSNR | SSIM | FSIM |
|---|---|---|---|---|---|---|---|---|
| | | 7 | 55 86 114 140 166 194 223 | 36 69 100 130 158 185 216 | 29 55 82 111 144 177 211 | 29.49377 | 0.86968 | 0.91845 |
| | APO | 4 | 76 117 160 206 | 74 115 158 207 | 82 121 163 204 | **25.27021** | 0.80767 | 0.85459 |
| | | 5 | 55 89 124 165 209 | 71 110 151 183 214 | 50 84 122 163 204 | 27.33734 | 0.83724 | 0.89500 |
| | | 6 | 55 88 121 155 188 220 | 64 99 128 156 184 214 | 45 82 112 146 180 213 | 28.04290 | 0.84830 | 0.90690 |
| | | 7 | 55 83 111 139 166 194 221 | 36 68 100 129 156 184 213 | 30 56 84 114 146 179 212 | 29.51470 | 0.86915 | 0.91772 |
| | GWO | 4 | 76 117 160 206 | 74 115 158 207 | 82 121 163 204 | **25.27021** | 0.80767 | 0.85459 |
| | | 5 | 55 89 124 165 209 | 72 111 151 183 214 | 51 84 122 163 204 | 27.35111 | **0.83866** | 0.89489 |
| | | 6 | 55 86 119 154 187 220 | 65 99 128 156 184 214 | 45 82 113 146 179 213 | 28.07804 | 0.84841 | 0.90850 |
| | | 7 | 55 84 113 140 166 194 223 | 36 68 101 129 156 184 214 | 29 55 83 113 146 179 213 | **29.51580** | **0.87018** | 0.91766 |
| | PSO | 4 | 76 117 160 206 | 74 115 158 207 | 82 121 163 204 | **25.27021** | 0.80767 | 0.85459 |
| | | 5 | 55 89 124 165 209 | 72 111 151 183 214 | 51 84 122 163 204 | 27.35111 | **0.83866** | 0.89489 |
| | | 6 | 55 86 119 154 187 220 | 65 99 128 156 184 214 | 45 82 113 146 179 213 | 28.07804 | 0.84841 | 0.90850 |
| | | 7 | 55 84 113 140 166 194 223 | 36 68 101 129 157 185 214 | 29 55 83 113 146 179 213 | 29.51334 | 0.87017 | 0.91732 |
| | DE | 4 | 76 117 160 206 | 74 115 158 207 | 82 122 163 204 | 25.26960 | **0.80771** | **0.85476** |
| | | 5 | 56 93 126 165 207 | 74 113 152 182 213 | 50 83 122 163 206 | 27.31723 | 0.83717 | **0.89764** |
| | | 6 | 56 85 117 154 190 221 | 63 93 126 154 184 213 | 42 82 114 149 182 215 | 27.97575 | 0.84527 | 0.90548 |
| | | 7 | 56 83 115 142 168 196 223 | 36 67 100 130 156 187 215 | 29 52 82 115 153 185 214 | 29.38661 | 0.86770 | 0.91808 |

**Table 8** (continued)

| Method | Algorithm | Thresholds | Red band | Green band | Blue band | PSNR | SSIM | FSIM |
|---|---|---|---|---|---|---|---|---|
| | MAQUATRE | 4 | 76 117 160 206 | 74 115 158 207 | 82 121 163 204 | **25.27021** | 0.80767 | 0.85459 |
| | | 5 | 55 89 124 165 209 | 72 111 151 183 214 | 51 84 122 163 204 | 27.35111 | **0.83866** | 0.89489 |
| | | 6 | 55 86 119 154 187 220 | 65 99 128 156 184 214 | 45 82 113 146 179 213 | 28.07804 | 0.84841 | 0.90850 |
| | | 7 | 55 84 113 140 166 194 223 | 36 68 101 129 156 184 213 | 29 56 84 115 149 182 214 | 29.51166 | 0.87005 | 0.91770 |
| | SGA | 4 | 76 117 160 206 | 74 115 158 207 | 82 121 163 204 | **25.27021** | 0.80767 | 0.85459 |
| | | 5 | 55 89 124 165 209 | 72 111 151 183 214 | 51 84 122 163 204 | 27.35111 | **0.83866** | 0.89489 |
| | | 6 | 55 87 120 154 187 220 | 65 99 128 156 184 214 | 45 82 113 146 179 213 | 28.07678 | 0.84888 | 0.90807 |
| | | 7 | 55 84 113 140 168 195 223 | 36 67 100 127 154 184 215 | 29 55 83 113 146 179 213 | 29.50570 | 0.86937 | **0.91857** |
| | AOO | 4 | 76 117 160 206 | 74 115 158 207 | 82 121 163 204 | **25.27021** | 0.80767 | 0.85459 |
| | | 5 | 55 89 124 165 209 | 72 111 151 183 214 | 51 84 122 163 204 | 27.35111 | **0.83866** | 0.89489 |
| | | 6 | 55 86 119 154 187 220 | 65 99 128 156 184 214 | 45 82 113 146 179 213 | 28.07804 | 0.84841 | 0.90850 |
| | | 7 | 55 84 113 140 166 194 223 | 36 67 100 128 156 185 215 | 29 55 83 113 146 179 213 | 29.50034 | 0.86938 | 0.91760 |

The best results are shown in bold

this, a new diversity metric (multidimensional volumes) is developed to guide population separation and merging processes. Additionally, a dynamic separation strategy is employed to improve search capability. Second, the autotrophic model in APO is refined by integrating a roulette-based fitness-distance balance method for selecting high-quality reference points to guide the population more effectively. To evaluate the performance of MAAPO, the algorithm is first tested on the CEC2017 test suite and compared against 12 state-of-the-art optimization algorithms, including APO, GWO, IGWO, PSO, PPSO, DE, ARO, FPA, GOA, MAQUATRE, SGA, and AOO. Subsequently, MAAPO is applied to multilevel thresholding image segmentation tasks. Experimental results demonstrate that MAAPO exhibits competitive performance compared to its counterparts. Specifically, it identifies superior solutions in global optimization, while in image segmentation, it delivers high-

**Table 9** The best thresholds of algorithms in the owl image

| Method | Algorithm | Thresholds | Red band | Green band | Blue band | PSNR | SSIM | FSIM |
|---|---|---|---|---|---|---|---|---|
| Otsu | MAAPO | 4 | 42 80 123 175 | 38 74 114 163 | 26 59 95 138 | **30.64616** | 0.94068 | 0.87811 |
| | | 5 | 37 67 101 139 186 | 34 63 94 129 172 | 22 49 78 110 148 | 32.16666 | **0.95331** | **0.89307** |
| | | 6 | 33 58 87 117 152 195 | 31 56 83 111 143 182 | 20 43 67 93 121 155 | **33.38550** | 0.96036 | 0.90556 |
| | | 7 | 31 51 75 102 129 160 200 | 28 49 72 96 122 153 188 | 18 38 59 80 103 129 159 | 34.46502 | **0.96589** | 0.91958 |
| | APO | 4 | 42 80 123 175 | 38 74 114 163 | 26 59 95 138 | **30.64616** | 0.94068 | 0.87811 |
| | | 5 | 37 68 102 139 186 | 34 63 94 129 172 | 22 49 77 109 147 | **32.16754** | 0.95329 | 0.89275 |
| | | 6 | 33 58 88 118 152 196 | 31 56 83 111 142 182 | 19 41 65 90 119 154 | 33.38410 | **0.96064** | 0.90537 |
| | | 7 | 29 48 73 100 129 160 200 | 30 52 74 98 125 155 191 | 18 38 60 82 105 132 163 | 34.46410 | 0.96550 | 0.92523 |
| | GWO | 4 | 42 80 123 175 | 38 74 114 163 | 26 59 95 138 | **30.64616** | 0.94068 | 0.87811 |
| | | 5 | 37 68 102 139 186 | 34 63 94 129 172 | 22 49 77 109 147 | **32.16754** | 0.95329 | 0.89275 |
| | | 6 | 33 58 87 117 152 195 | 31 56 83 111 143 182 | 20 43 67 92 120 154 | 33.38543 | 0.96038 | 0.90552 |
| | | 7 | 30 50 75 101 129 161 201 | 29 50 73 97 123 153 189 | 18 38 59 81 105 132 163 | **34.47889** | 0.96572 | 0.91904 |
| | PSO | 4 | 42 80 123 175 | 38 74 114 163 | 26 59 95 138 | **30.64616** | 0.94068 | 0.87811 |
| | | 5 | 37 68 102 139 186 | 34 63 94 129 172 | 22 49 77 109 147 | **32.16754** | 0.95329 | 0.89275 |
| | | 6 | 33 58 87 117 152 195 | 31 56 83 111 143 182 | 20 43 67 92 120 154 | 33.38543 | 0.96038 | 0.90552 |
| | | 7 | 30 50 75 101 129 161 201 | 29 50 73 97 123 153 189 | 18 38 59 81 105 132 163 | **34.47889** | 0.96572 | 0.91904 |
| | DE | 4 | 42 80 122 174 | 38 74 114 163 | 26 59 95 139 | 30.64542 | **0.94075** | **0.87834** |
| | | 5 | 38 67 102 140 189 | 34 65 96 130 173 | 22 49 78 109 149 | 32.15533 | 0.95309 | 0.89210 |
| | | 6 | 32 57 84 118 151 194 | 32 58 85 111 145 186 | 19 40 63 88 116 153 | 33.35487 | 0.96012 | **0.90806** |

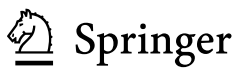

**Table 9** (continued)

| Method | Algorithm | Thresholds | Red band | Green band | Blue band | PSNR | SSIM | FSIM |
|---|---|---|---|---|---|---|---|---|
| | | 7 | 27 46 72 97 127 160 202 | 28 49 69 93 123 152 191 | 16 37 58 80 102 127 161 | 34.40190 | 0.96412 | **0.94895** |
| | MAQUATRE | 4 | 42 80 123 175 | 38 74 114 163 | 26 59 95 138 | **30.64616** | 0.94068 | 0.87811 |
| | | 5 | 37 68 102 139 186 | 34 63 94 129 172 | 22 49 77 109 147 | **32.16754** | 0.95329 | 0.89275 |
| | | 6 | 33 58 87 117 152 195 | 31 56 83 111 143 182 | 20 43 67 92 120 154 | 33.38543 | 0.96038 | 0.90552 |
| | | 7 | 30 50 75 101 129 161 201 | 29 50 72 96 122 152 189 | 18 38 59 81 105 131 162 | 34.47708 | 0.96574 | 0.91881 |
| | SGA | 4 | 42 80 123 175 | 38 74 114 163 | 26 59 95 138 | **30.64616** | 0.94068 | 0.87811 |
| | | 5 | 37 68 102 139 186 | 34 63 94 129 172 | 22 49 77 109 147 | **32.16754** | 0.95329 | 0.89275 |
| | | 6 | 33 58 87 117 151 194 | 31 56 82 110 142 182 | 20 43 67 92 120 154 | 33.38450 | 0.96042 | 0.90541 |
| | | 7 | 30 50 76 103 130 162 202 | 29 50 73 97 123 153 189 | 18 37 58 80 104 131 162 | 34.47666 | 0.96575 | 0.91876 |
| | AOO | 4 | 42 80 123 175 | 38 74 114 163 | 26 59 95 138 | **30.64616** | 0.94068 | 0.87811 |
| | | 5 | 37 68 102 139 186 | 34 63 94 129 172 | 22 49 77 109 147 | **32.16754** | 0.95329 | 0.89275 |
| | | 6 | 33 58 87 117 152 195 | 31 56 83 111 143 182 | 20 43 67 92 120 154 | 33.38543 | 0.96038 | 0.90552 |
| | | 7 | 30 50 75 101 129 162 202 | 29 50 73 96 122 152 189 | 18 38 59 81 105 132 163 | 34.47666 | 0.96575 | 0.91876 |
| Kapur | MAAPO | 4 | 42 96 147 199 | 37 92 145 193 | 29 92 152 231 | 29.10720 | 0.92056 | 0.87744 |
| | | 5 | 39 84 126 169 210 | 32 77 122 166 212 | 21 70 119 168 231 | 30.56550 | 0.93618 | 0.89038 |
| | | 6 | 37 71 108 143 178 216 | 31 70 109 147 182 217 | 19 57 93 130 170 231 | **32.10104** | **0.94920** | 0.89974 |
| | | 7 | 37 68 99 131 163 193 222 | 29 63 96 128 159 191 221 | 15 46 81 115 151 186 231 | 33.03322 | 0.95570 | 0.90623 |
| | APO | 4 | 42 96 147 199 | 37 92 145 193 | 29 92 152 231 | 29.10720 | 0.92056 | 0.87744 |
| | | 5 | 39 84 127 169 210 | 32 78 122 166 212 | 21 70 118 168 231 | 30.55628 | 0.93600 | 0.89029 |

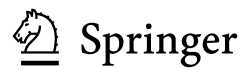

**Table 9** (continued)

| Method | Algorithm | Thresholds | Red band | Green band | Blue band | PSNR | SSIM | FSIM |
|---|---|---|---|---|---|---|---|---|
| | | 6 | 38 73 108 145 180 216 | 31 70 108 146 183 217 | 17 54 93 131 170 231 | 32.04944 | 0.94908 | 0.89868 |
| | | 7 | 36 67 97 127 160 192 222 | 29 60 91 125 158 191 222 | 16 51 85 116 151 186 231 | 33.10755 | **0.95607** | 0.90670 |
| | GWO | 4 | 42 96 147 199 | 37 92 145 193 | 29 92 152 231 | 29.10720 | 0.92056 | 0.87744 |
| | | 5 | 39 83 126 169 210 | 32 78 122 166 212 | 21 70 118 168 231 | **30.57499** | **0.93619** | **0.89080** |
| | | 6 | 38 74 109 145 180 216 | 31 70 108 146 182 217 | 16 55 93 131 170 231 | 31.99765 | 0.94800 | 0.89895 |
| | | 7 | 36 67 99 131 161 192 222 | 29 62 95 127 159 191 221 | 15 48 83 118 152 188 231 | 33.01529 | 0.95535 | 0.90715 |
| | PSO | 4 | 42 96 147 199 | 37 92 145 193 | 29 92 152 231 | 29.10720 | 0.92056 | 0.87744 |
| | | 5 | 39 83 126 169 210 | 32 78 122 166 212 | 21 70 118 168 231 | **30.57499** | **0.93619** | **0.89080** |
| | | 6 | 38 74 109 145 180 216 | 31 70 108 146 182 217 | 17 55 93 131 170 231 | 32.03572 | 0.94857 | 0.89842 |
| | | 7 | 36 67 99 131 161 192 222 | 29 62 95 127 159 191 221 | 15 50 84 118 152 188 231 | 32.99415 | 0.95487 | 0.90727 |
| | DE | 4 | 42 96 147 199 | 38 92 145 193 | 29 92 152 231 | **29.12697** | **0.92076** | **0.87770** |
| | | 5 | 39 85 127 172 212 | 32 77 122 166 211 | 22 73 122 168 231 | 30.50238 | 0.93523 | 0.88976 |
| | | 6 | 38 75 112 146 181 219 | 29 72 110 145 182 215 | 18 57 96 136 172 231 | 31.85058 | 0.94657 | **0.89994** |
| | | 7 | 38 67 99 134 163 192 224 | 30 64 93 122 157 191 220 | 14 56 92 127 163 193 231 | 32.65446 | 0.95155 | 0.90543 |
| | MAQUATRE | 4 | 42 96 147 199 | 37 92 145 193 | 29 92 152 231 | 29.10720 | 0.92056 | 0.87744 |
| | | 5 | 39 83 126 169 210 | 32 78 122 166 212 | 21 70 118 168 231 | **30.57499** | **0.93619** | **0.89080** |
| | | 6 | 38 74 109 145 180 216 | 31 70 108 146 182 217 | 17 55 93 131 170 231 | 32.03572 | 0.94857 | 0.89842 |
| | | 7 | 35 64 95 126 158 191 222 | 29 61 93 125 157 188 221 | 16 51 85 118 152 188 231 | **33.12053** | 0.95604 | **0.90852** |
| | SGA | 4 | 42 96 147 199 | 37 92 145 193 | 29 92 152 231 | 29.10720 | 0.92056 | 0.87744 |

**Table 9** (continued)

| Method | Algorithm | Thresholds | Red band | Green band | Blue band | PSNR | SSIM | FSIM |
|---|---|---|---|---|---|---|---|---|
| | | 5 | 39 83 126 169 210 | 32 78 122 166 212 | 21 70 118 168 231 | **30.57499** | **0.93619** | **0.89080** |
| | | 6 | 38 74 109 145 180 216 | 31 70 108 146 182 217 | 17 56 95 134 170 231 | 31.99883 | 0.94825 | 0.89850 |
| | | 7 | 36 67 99 131 161 192 222 | 29 61 93 125 158 191 221 | 15 49 84 119 152 186 231 | 33.03698 | 0.95547 | 0.90714 |
| | AOO | 4 | 42 96 147 199 | 37 92 145 193 | 29 92 152 231 | 29.10720 | 0.92056 | 0.87744 |
| | | 5 | 39 83 126 169 210 | 32 78 122 166 212 | 21 70 118 168 231 | **30.57499** | **0.93619** | **0.89080** |
| | | 6 | 38 74 109 145 180 216 | 31 70 108 146 182 217 | 17 55 93 131 170 231 | 31.99883 | 0.94825 | 0.89850 |
| | | 7 | 36 67 99 131 161 192 222 | 29 62 95 127 159 191 221 | 15 50 84 117 152 188 231 | 32.99916 | 0.95490 | 0.90728 |

quality segmented images as measured by PSNR, SSIM, and FSIM metrics. The success of MAAPO provides two key insights for enhancing algorithmic performance in the evolutionary computing community. First, the separating-merging principle derived from membrane computing can be employed during the optimization process to dynamically adjust population size, thereby achieving a better balance between exploration and exploitation. Second, integrate existing selection methods or design new mechanisms to effectively guide the population by identifying promising reference points. These strategies are not limited to MAAPO and hold potential for improving a wide range of meta-heuristic algorithms. In the domain of image processing, this study proposes a new image segmentation approach based on a meta-heuristic algorithm. For multilevel thresholding image segmentation, the algorithm uses Otsu's method and Kapur's entropy as separate objective functions to guide the search for optimal thresholds. The results validate its performance and general applicability.

Despite the promising results achieved in this study, several limitations should be noted. First, the added improvements slightly increase the runtime compared to APO. Second, only the Otsu method and Kapur entropy were used for image segmentation, while other methods like Cross-entropy and Renyi entropy remain unexplored. Finally, the experiments were limited to popular animal images. Testing the algorithm on more diverse datasets, such as those from medical, remote sensing, industrial, or traffic domains, could better demonstrate its robustness.

Future studies may build upon this work in several ways. First, the proposed MAAPO algorithm can be applied to various practical engineering optimization tasks. Furthermore, the algorithm can be extended to tackle high-dimensional, multi-objective, and binary optimization problems. To develop the surrogate algorithm, it is also possible to integrate several machine learning models. Last but not least, hybridization with other optimizers may be explored to enhance overall performance.

**Table 10** The best thresholds of algorithms in the horse image

| Method | Algorithm | Thresholds | Red band | Green band | Blue band | PSNR | SSIM | FSIM |
|---|---|---|---|---|---|---|---|---|
| Otsu | MAAPO | 4 | 89 131 166 211 | 88 135 176 216 | 66 93 117 177 | 27.58414 | 0.81675 | 0.89485 |
| | | 5 | 84 121 149 181 220 | 85 128 162 189 224 | 60 84 103 125 186 | **29.27343** | 0.85456 | **0.92563** |
| | | 6 | 82 116 139 163 192 226 | 82 122 154 176 201 231 | 57 78 96 111 132 191 | 30.55765 | 0.88085 | 0.94150 |
| | | 7 | 76 107 128 147 172 198 229 | 76 106 135 159 178 202 231 | 49 66 85 100 114 135 191 | 31.67025 | 0.89826 | **0.95621** |
| | APO | 4 | 89 131 166 211 | 88 135 176 216 | 66 93 117 177 | 27.58414 | 0.81675 | 0.89485 |
| | | 5 | 84 121 149 181 220 | 84 128 162 189 224 | 60 84 103 125 186 | 29.27309 | 0.85448 | 0.92509 |
| | | 6 | 82 117 140 165 194 227 | 82 121 152 174 199 230 | 57 78 96 111 132 190 | 30.55427 | 0.88023 | **0.94173** |
| | | 7 | 77 108 130 150 173 198 229 | 77 108 138 161 180 203 232 | 48 66 84 99 114 136 195 | 31.67638 | 0.89771 | 0.95577 |
| | GWO | 4 | 89 131 166 211 | 88 135 176 216 | 66 93 117 177 | 27.58414 | 0.81675 | 0.89485 |
| | | 5 | 84 121 149 181 220 | 85 128 162 189 224 | 60 84 103 125 186 | **29.27343** | 0.85456 | **0.92563** |
| | | 6 | 82 116 139 163 192 226 | 82 122 153 174 199 230 | 57 78 96 111 132 191 | **30.55907** | 0.88083 | 0.94135 |
| | | 7 | 77 107 128 148 171 198 229 | 78 110 139 161 180 204 233 | 49 67 85 100 114 135 193 | 31.68718 | 0.89842 | 0.95567 |
| | PSO | 4 | 89 131 166 211 | 88 135 176 216 | 66 93 117 177 | 27.58414 | 0.81675 | 0.89485 |
| | | 5 | 84 121 149 181 220 | 85 128 162 189 224 | 60 84 103 125 186 | **29.27343** | 0.85456 | **0.92563** |
| | | 6 | 82 116 139 163 192 226 | 82 122 153 174 199 230 | 57 78 96 111 132 191 | **30.55907** | 0.88083 | 0.94135 |
| | | 7 | 77 107 128 148 171 198 229 | 78 110 139 161 180 204 233 | 48 66 84 99 113 134 192 | **31.68746** | **0.89879** | 0.95574 |
| | DE | 4 | 90 132 167 212 | 89 135 175 216 | 66 93 117 176 | **27.58480** | **0.81692** | **0.89504** |
| | | 5 | 84 120 147 179 220 | 82 126 162 190 224 | 60 83 102 124 187 | 29.25549 | **0.85501** | 0.92388 |

**Table 10** (continued)

| Method | Algorithm | Thresholds | Red band | Green band | Blue band | PSNR | SSIM | FSIM |
|---|---|---|---|---|---|---|---|---|
| | | 6 | 81 115 139 161 192 224 | 84 126 156 177 201 234 | 59 81 98 112 135 191 | 30.51296 | **0.88127** | 0.94150 |
| | | 7 | 75 104 125 146 168 192 225 | 76 108 135 159 180 202 230 | 49 67 85 100 113 136 189 | 31.62325 | 0.89720 | 0.95505 |
| | MAQUATRE | 4 | 89 131 166 211 | 88 135 176 216 | 66 93 117 177 | 27.58414 | 0.81675 | 0.89485 |
| | | 5 | 84 121 149 181 220 | 85 128 162 189 224 | 60 84 103 125 186 | **29.27343** | 0.85456 | **0.92563** |
| | | 6 | 82 116 139 163 192 226 | 82 122 153 174 199 230 | 57 78 96 111 132 191 | **30.55907** | 0.88083 | 0.94135 |
| | | 7 | 77 107 128 148 171 198 229 | 78 108 137 160 180 204 233 | 48 66 84 99 113 134 191 | 31.68409 | 0.89837 | 0.95596 |
| | SGA | 4 | 89 131 166 211 | 88 135 176 216 | 66 93 117 177 | 27.58414 | 0.81675 | 0.89485 |
| | | 5 | 84 121 149 181 220 | 85 128 162 189 224 | 60 84 103 125 186 | **29.27343** | 0.85456 | **0.92563** |
| | | 6 | 82 116 139 163 192 226 | 83 122 153 174 199 230 | 57 78 96 111 132 191 | **30.55907** | 0.88085 | 0.94153 |
| | | 7 | 77 107 128 148 171 198 230 | 78 110 139 161 180 204 233 | 47 65 84 99 114 136 194 | 31.68299 | 0.89814 | 0.95601 |
| | AOO | 4 | 89 131 166 211 | 88 135 176 216 | 66 93 117 177 | 27.58414 | 0.81675 | 0.89485 |
| | | 5 | 84 121 149 181 220 | 85 128 162 189 224 | 60 84 103 125 186 | **29.27343** | 0.85456 | **0.92563** |
| | | 6 | 82 116 139 163 192 226 | 82 122 153 174 199 230 | 57 78 96 111 132 191 | **30.55907** | 0.88083 | 0.94135 |
| | | 7 | 77 107 128 148 171 198 229 | 77 109 138 160 180 204 233 | 48 66 84 99 113 134 193 | 31.68676 | 0.89828 | 0.95585 |
| Kapur | MAAPO | 4 | 28 103 154 201 | 37 93 140 192 | 34 82 132 174 | 25.09489 | 0.74493 | 0.83868 |
| | | 5 | 28 99 140 176 214 | 37 93 140 179 213 | 34 81 122 151 180 | 26.94422 | 0.79699 | **0.88670** |
| | | 6 | 28 70 104 142 179 216 | 37 77 109 142 179 211 | 36 81 122 154 186 225 | **27.74821** | 0.80942 | **0.91592** |

**Table 10** (continued)

| Method | Algorithm | Thresholds | Red band | Green band | Blue band | PSNR | SSIM | FSIM |
|---|---|---|---|---|---|---|---|---|
| | | 7 | 28 69 99 128 159 191 221 | 37 78 109 140 168 196 223 | 33 63 89 123 153 186 225 | **29.68205** | 0.84895 | 0.93917 |
| | APO | 4 | 28 103 154 201 | 37 93 140 192 | 34 82 132 174 | 25.09489 | 0.74493 | 0.83868 |
| | | 5 | 28 100 140 176 213 | 37 92 140 179 213 | 34 81 122 151 180 | 26.92474 | 0.79677 | 0.88550 |
| | | 6 | 28 69 103 142 179 215 | 37 78 110 143 179 213 | 34 81 123 154 186 225 | 27.71833 | 0.80829 | 0.91585 |
| | | 7 | 28 69 100 131 160 190 221 | 37 77 109 139 168 194 220 | 32 62 89 123 154 186 223 | 29.65713 | **0.84919** | 0.93779 |
| | GWO | 4 | 28 103 154 201 | 37 93 140 192 | 34 82 132 174 | 25.09489 | 0.74493 | 0.83868 |
| | | 5 | 28 99 140 176 213 | 37 93 140 179 213 | 34 81 122 151 180 | 26.93921 | **0.79705** | 0.88616 |
| | | 6 | 28 70 103 142 179 215 | 37 77 109 142 179 213 | 34 81 122 154 186 225 | 27.73601 | 0.80926 | 0.91491 |
| | | 7 | 28 69 100 130 160 191 222 | 37 77 108 140 169 195 222 | 32 63 90 124 154 186 225 | 29.68134 | 0.84821 | **0.93946** |
| | PSO | 4 | 28 103 154 201 | 37 93 140 192 | 34 82 132 174 | 25.09489 | 0.74493 | 0.83868 |
| | | 5 | 28 99 140 176 213 | 37 93 140 179 213 | 34 81 122 151 180 | 26.93921 | **0.79705** | 0.88616 |
| | | 6 | 28 70 103 142 179 215 | 37 77 109 142 179 213 | 34 81 122 154 186 225 | 27.73601 | 0.80926 | 0.91491 |
| | | 7 | 28 69 100 130 160 191 222 | 37 77 108 140 169 195 222 | 32 62 89 124 154 186 225 | 29.64343 | 0.84713 | 0.93934 |
| | DE | 4 | 28 103 154 201 | 37 93 140 192 | 34 82 131 174 | **25.10974** | **0.74556** | **0.83878** |
| | | 5 | 28 100 140 179 215 | 37 91 139 179 214 | 34 81 121 152 182 | **26.94629** | 0.79615 | 0.88637 |
| | | 6 | 28 70 102 141 178 215 | 37 77 108 140 179 212 | 36 80 122 153 186 223 | 27.71833 | 0.80923 | 0.91468 |
| | | 7 | 28 67 98 128 160 191 219 | 37 79 108 139 174 200 227 | 31 61 90 121 153 186 226 | 29.60106 | 0.84781 | 0.93791 |
| | MAQUATRE | 4 | 28 103 154 201 | 37 93 140 192 | 34 82 132 174 | 25.09489 | 0.74493 | 0.83868 |

**Table 10** (continued)

| Method | Algorithm | Thresholds | Red band | Green band | Blue band | PSNR | SSIM | FSIM |
|---|---|---|---|---|---|---|---|---|
| | | 5 | 28 99 140 176 213 | 37 93 140 179 213 | 34 81 122 151 180 | 26.93921 | **0.79705** | 0.88616 |
| | | 6 | 28 70 103 142 179 215 | 37 77 109 142 179 213 | 34 81 122 154 186 225 | 27.73601 | 0.80926 | 0.91491 |
| | | 7 | 28 69 100 131 162 193 223 | 37 77 108 140 170 196 222 | 32 62 89 124 154 186 225 | 29.60289 | 0.84659 | 0.93869 |
| | SGA | 4 | 28 103 154 201 | 37 93 140 192 | 34 82 132 174 | 25.09489 | 0.74493 | 0.83868 |
| | | 5 | 28 100 141 178 215 | 37 93 140 179 213 | 34 81 122 151 180 | 26.91634 | 0.79605 | 0.88607 |
| | | 6 | 28 70 103 142 178 214 | 37 78 110 142 179 213 | 34 81 122 154 186 225 | 27.74087 | **0.80983** | 0.91493 |
| | | 7 | 28 69 100 131 162 192 223 | 37 77 108 140 171 196 223 | 32 62 89 123 154 186 225 | 29.64513 | 0.84819 | 0.93897 |
| | AOO | 4 | 28 103 154 201 | 37 93 140 192 | 34 82 132 174 | 25.09489 | 0.74493 | 0.83868 |
| | | 5 | 28 99 140 176 213 | 37 93 140 179 213 | 34 81 122 151 180 | 26.93921 | **0.79705** | 0.88616 |
| | | 6 | 28 70 103 142 179 215 | 37 77 109 142 179 213 | 34 81 122 154 186 225 | 27.73601 | 0.80926 | 0.91491 |
| | | 7 | 28 69 100 130 160 191 222 | 37 76 108 140 169 195 222 | 32 62 89 124 154 186 225 | 29.64193 | 0.84709 | 0.93914 |

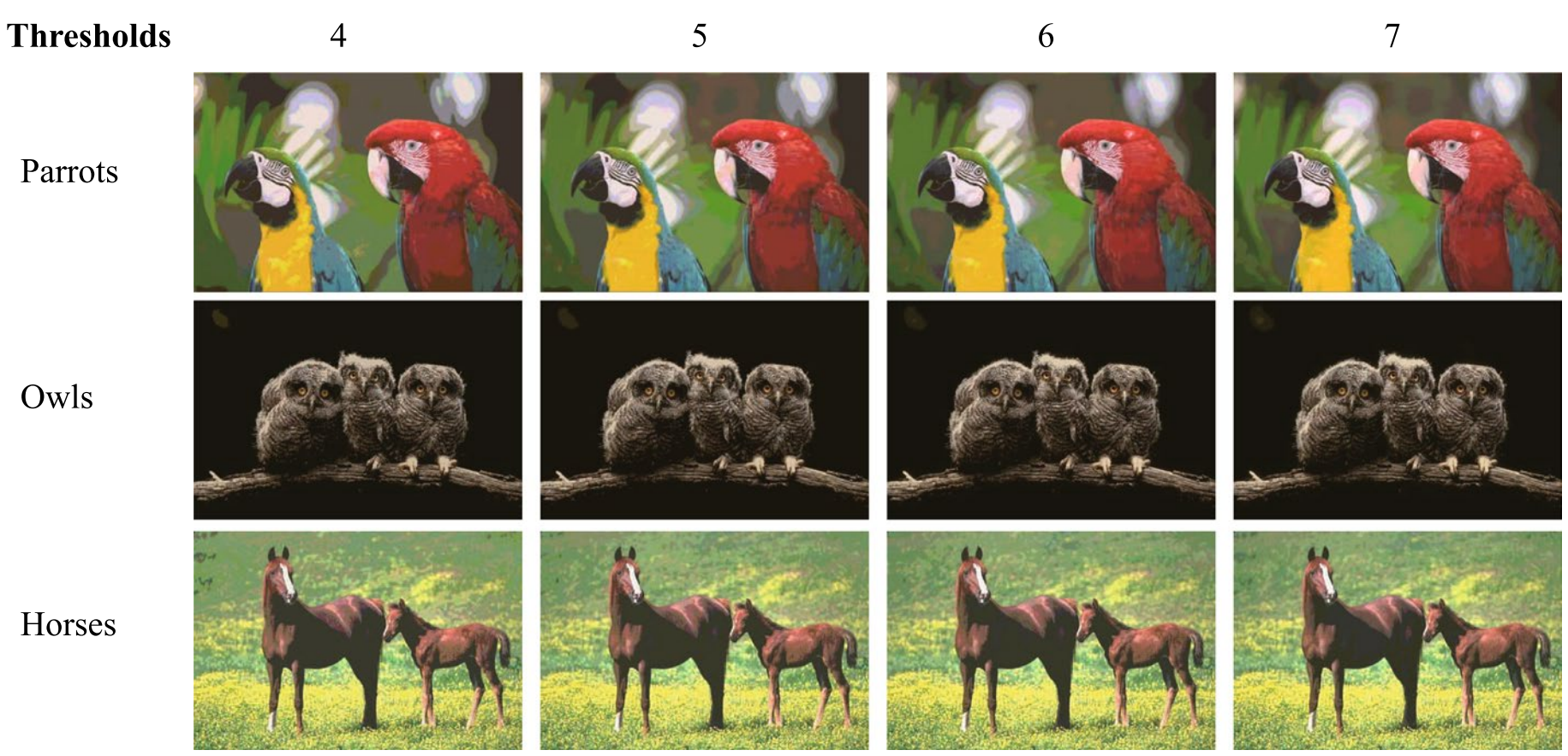

**Fig. 8** The segmented images of MAAPO using the Otsu method

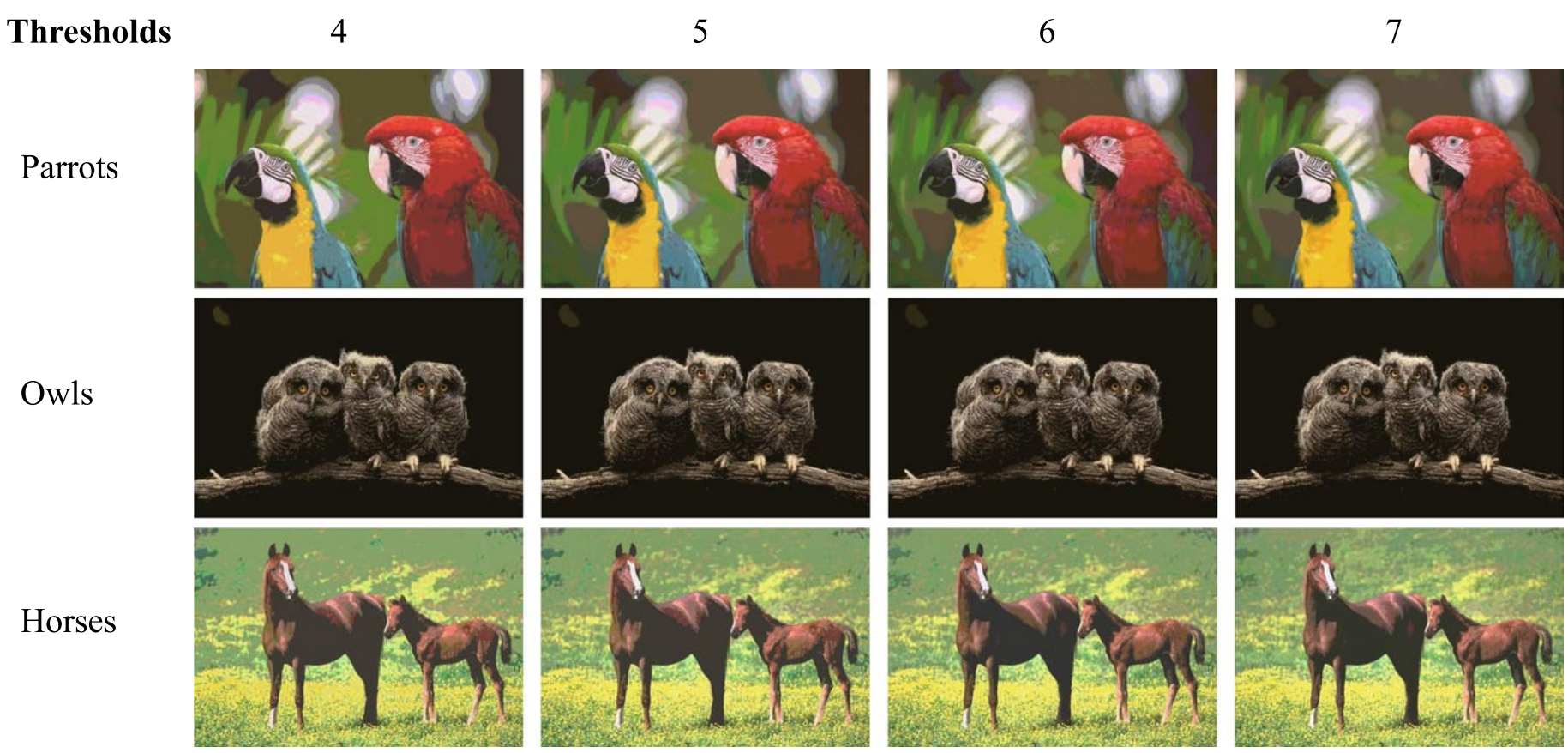

**Fig. 9** The segmented images of MAAPO using the Kapur entropy method

**Table 11** The average ranking of algorithms using the Otsu method

| Metric | MAAPO | APO | GWO | PSO | DE | MAQUATRE | SGA | AOO |
|---|---|---|---|---|---|---|---|---|
| PSNR | 5.00 | 5.88 | 3.25 | **3.17** | 7.42 | 3.54 | 4.17 | 3.58 |
| SSIM | **3.83** | 6.04 | 4.38 | 4.29 | 4.50 | 4.33 | 4.33 | 4.29 |
| FSIM | 3.79 | 4.46 | 4.88 | 4.79 | **3.58** | 4.67 | 4.96 | 4.88 |

**Table 12** The average ranking of algorithms using the Kapur entropy method

| Metric | MAAPO | APO | GWO | PSO | DE | MAQUATRE | SGA | AOO |
|---|---|---|---|---|---|---|---|---|
| PSNR | **3.17** | 4.83 | 4.04 | 4.33 | 6.25 | 4.08 | 4.54 | 4.75 |
| SSIM | **3.75** | 5.08 | 4.04 | 4.25 | 5.83 | 4.08 | 4.29 | 4.67 |
| FSIM | **3.42** | 5.25 | 4.29 | 4.83 | 4.08 | 4.75 | 4.71 | 4.67 |

The best results are shown in bold

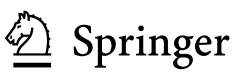

**Supplementary Information** The online version contains supplementary material available at https://doi.org/10.1007/s10462-025-11319-2.

**Acknowledgements** The authors gratefully acknowledge financial support CZ.02.01.01/00/22-008/0004590 by the Czech Republic Ministry of Education, Youth, and Sports in the project "Robotics and Advanced Industrial Production" (ROBOPROX).

**Author Contributions** Xiaopeng Wang: Methodology, Software, Visualization, Writing-original draft. Václav Snášel: Conceptualization, Methodology, Writing-review & editing. Seyedali Mirjalili: Visualization, Writing-review & editing. Jeng-Shyang Pan: Methodology, Writing-review & editing.

**Funding** Open access publishing supported by the institutions participating in the CzechELib Transformative Agreement.

## Declarations

**Conflict of interest** The authors declare that they have no known competing financial interests or personal relationships that may have influenced the work reported in this paper.